\documentclass[twocolumn]{USG}

\usepackage{anyfontsize}
\usepackage{colortbl}   % For coloring table lines
\usepackage{xcolor}     % For color definitions
\usepackage{steinmetz}
\usepackage{amsmath,amssymb,amsfonts}
\usepackage{booktabs}
\usepackage{tabularx}
\usepackage{multirow}
\usepackage{url}
\usepackage{algorithm}
\usepackage{algpseudocode}
\usepackage{graphicx}
\usepackage{placeins}
\usepackage{flafter}
\usepackage{float}

\graphicspath{{./images/}}

\articletype{RESEARCH ARTICLE}%
\journal{Advanced Intelligent Systems}
\copyyear{2026}
\articledoi{}

\begin{document}

% Prevent excessive vertical stretching between paragraphs and section headings
% in the two-column layout while preserving the journal class's native spacing.
\raggedbottom

% ---------------- Article Header ----------------
\title{FD-CanKD: Frequency-Decoupled Cross-Attention\protect\\
Distillation as a Refinement Prior for\protect\\
Compact Object Detectors}
% \subtitle{}
% \transtitle{}
% \subtranstitle{}

\author[1]{YoungJae Cheong}
\author[1]{Jhonghyun An}

\authormark{CHEONG and AN}
\titlemark{FD-CanKD for YOLOv12}

\address[1]{\orgdiv{Department of AI and Software}, \orgname{Gachon University}, %
\orgaddress{\city{Seongnam}, \country{Republic of Korea}}}

\corres{Jhonghyun An (\email{jhonghyun@gachon.ac.kr})}

\fundingInfo{This work was supported by the Institute of Information \& Communications Technology Planning \& Evaluation (IITP) grant funded by the Ministry of Science and ICT (MSIT), Republic of Korea (No. RS-2024-00397615).}

\keywords{Object detection | YOLOv12 | knowledge distillation | cross-attention distillation | frequency-decoupled distillation}

\abstract[ABSTRACT]{Compact object detectors are suitable for resource-constrained visual perception, but their limited representation capacity creates an accuracy gap relative to large models. Conventional detector distillation often relies on prediction-level supervision or a single feature-alignment target, such as response, distribution, correlation, or frequency-domain matching. Frequency-Decoupled Cross-Attention Knowledge Distillation (FD-CanKD) is presented as a detector-oriented framework that transfers teacher knowledge at three complementary levels: head-level prediction supervision, relation-level non-local context transfer, and frequency-level component-selective alignment. Student features first aggregate teacher-side spatial context through cross-attention-based relation transfer, after which frequency-aware alignment preserves complementary structural and detail-sensitive cues. Under controlled Microsoft Common Objects in Context (COCO) experiments, fixed 50-epoch from-scratch comparisons show that FD-CanKD remains competitive with representative detector knowledge distillation baselines. Post-distillation continued fine-tuning further produces a stronger refinement-ready student than detector-only fine-tuning, reaching 48.87 mean average precision (mAP) at intersection-over-union thresholds from 0.50 to 0.95 (\(\mathrm{mAP}_{50:95}\)), 65.84 \(\mathrm{mAP}_{50}\), and 53.40 \(\mathrm{mAP}_{75}\) after 20 additional epochs. All distillation modules are removed after training, leaving the deployed student unchanged at \(19.7\mathrm{M}\) parameters. The framework is instantiated and evaluated in a controlled YOLOv12 teacher--student setting as a representative compact-detector case study.}

\abbr{COCO, Common Objects in Context. DFL, Distribution Focal Loss. FD-CanKD, Frequency-Decoupled Cross-Attention Knowledge Distillation. FFT, fast Fourier transform. KD, Knowledge Distillation. mAP, mean average precision. MSE, mean squared error. YOLO, You Only Look Once.}

\maketitle

% ---------------- Introduction ----------------
\section{Introduction}

Object detection is a fundamental task for autonomous driving, surveillance, robotics, remote sensing, and edge-based visual perception. Among efficient one-stage detectors, the You Only Look Once (YOLO) family has been widely adopted because it provides a practical balance between detection accuracy and
inference efficiency. Since the original YOLO, subsequent variants have improved multi-scale prediction, feature representation, training strategies, deployment efficiency, and end-to-end detection
capability~\cite{redmon2016yolo,redmon2017yolo9000,redmon2018yolov3,bochkovskiy2020yolov4,jocher2020yolov5,li2023yolov6,wang2023yolov7,jocher2023yolov8,wang2024yolov9,wang2024yolov10,jocher2024yolov11}. Recently, YOLOv12 introduced an attention-centric efficient detector that enhances representation quality while maintaining efficient inference~\cite{tian2025yolov12}. Beyond YOLO-based detectors,
DETR-style efficient detectors have also improved query denoising, encoder efficiency, and end-to-end detection performance~\cite{li2022dndetr,zhao2024rtdetr,lv2024rtdetrv2}.

Despite these advances, a clear accuracy and efficiency gap remains between large-scale and compact detectors. Large-scale detectors generally provide stronger representation capacity and higher detection accuracy, but they require considerable computation and memory. Compact detectors are more suitable for resource-constrained deployment, but their limited capacity often leads to degraded classification confidence, localization accuracy, and localized detection responses. Therefore, transferring informative
representations from a high-capacity teacher detector to a compact student detector is an important problem for accurate and efficient object detection, regardless of the particular detector family used
to instantiate the teacher and student.

Knowledge distillation (KD) provides a promising solution by transferring knowledge from a teacher model to a student model~\cite{bucilua2006model,hinton2015distilling}. Early studies mainly focused on output-level distillation and feature hints, while later methods improved the transfer process through mutual learning, decoupled logits, output correlations, hierarchical feature review, relational constraints, correlation-based alignment, and selective feature imitation~\cite{romero2014fitnets,zhang2018dml,
zhao2022dkd,huang2022dist,chen2021review,park2019rkd,cao2022pkd,heo2019overhaul}. For object detection, distillation has further evolved toward detector-specific supervision, including intermediate feature transfer, foreground and global feature distillation, channel-wise response alignment, masked generative
feature reconstruction, diffusion-based feature refinement, instance-conditional supervision, distillation search, kernel alignment, and frequency-aware transfer~\cite{li2017mimic,wang2021mimicking,zhang2020fkd,yang2022fgd,shu2021cwd,yang2022mgd,huang2023diffkd,kang2021icd,li2024detkds,zhou2024pcka,zhang2024freekd}. Despite this broad progress, direct target matching remains a common baseline and may become restrictive when heterogeneous teacher responses are transferred to a substantially smaller student using a single alignment rule.

\begin{figure}[!t]
  \centering
  \includegraphics[width=\columnwidth]{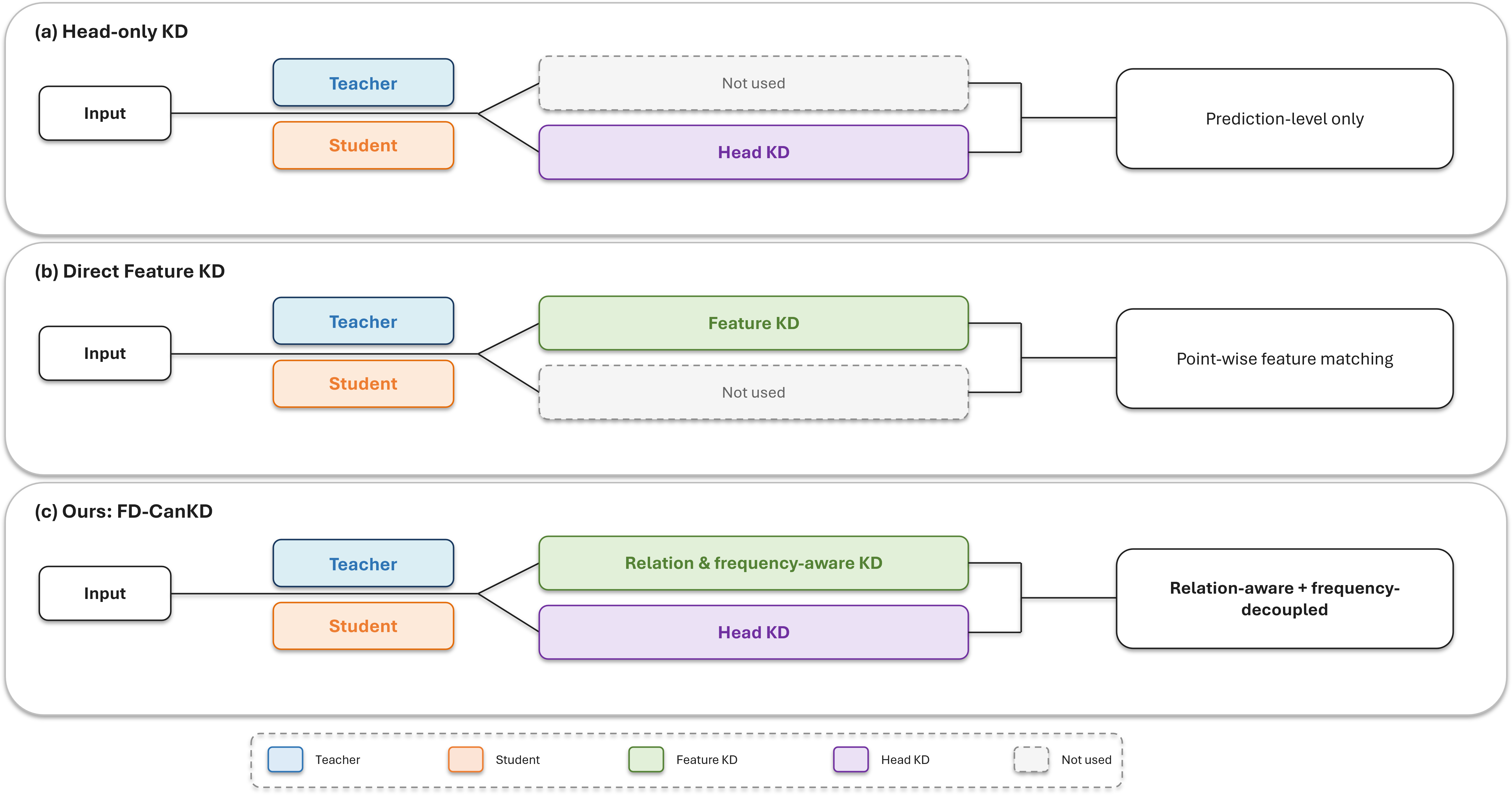}
  \caption{\textbf{Comparison of knowledge distillation paradigms for YOLOv12 large-to-small distillation.} (a) Head-only knowledge distillation (KD) provides prediction-level supervision only. (b) Direct feature KD performs point-wise feature matching between the teacher and student. (c) FD-CanKD combines head KD, relation-aware transfer, and band-specific feature alignment within a detector-oriented training pipeline~\cite{sun2025cankd,liu2025fdcmkd}.}
  \label{fig:kd_comparison}
\end{figure}

As illustrated in Figure~\ref{fig:kd_comparison}, head-only distillation transfers the final prediction behavior of the teacher, but it does not explicitly guide the intermediate feature hierarchy. Direct feature distillation transfers feature-level information, but it commonly relies on point-wise matching between teacher and student features. Such rigid matching can overlook non-local teacher-student spatial relations and can make the student follow teacher-specific local responses. This issue is particularly important for compact detectors because teacher features contain heterogeneous information, including object-level structures, contextual responses, boundary-sensitive activations, and local variations that may not be equally transferable. In contrast, Figure~\ref{fig:kd_comparison}(c) summarizes the detector-oriented integration adopted in FD-CanKD, where head-output supervision is coupled with relation-aware feature transfer and component-selective alignment.

% ---------------- Figure 1 ----------------

These observations indicate that large-to-small detector distillation should not simply increase the amount of teacher supervision. Instead, teacher knowledge should be organized according to its role
in dense object detection. Prediction outputs should guide the final classification and localization behavior of the student, multi-scale feature relations should transfer contextual dependencies among
objects and surrounding regions, and a component-selective alignment mechanism should avoid imposing the same matching strength on all transferred responses. In this study, frequency decomposition is used
as one practical mechanism for implementing such selective alignment rather than being treated as a general deficiency of prior detector distillation methods.

Attention and relation modeling provide a natural mechanism for transferring contextual feature dependencies. Transformer attention captures long-range dependencies~\cite{vaswani2017attention},
non-local neural networks model global spatial relations beyond local convolutional neighborhoods~\cite{wang2018nonlocal}, and relation networks exploit contextual interactions for object detection~\cite{hu2018relation}. Based on this principle, cross-attention-based non-local feature transfer allows student features to aggregate teacher-side spatial context~\cite{sun2025cankd}. In FD-CanKD, this relation transfer is applied to semantically corresponding multi-scale detection features, allowing the student to refer to teacher responses across spatial positions rather than matching only
the same spatial location. Channel and resolution differences between the selected teacher and student features can be handled through learnable projections and spatial resampling.

Frequency-domain analysis provides one useful perspective for partitioning heterogeneous detection responses. Prior studies on cross-modal distillation and feature distribution alignment show that
representation inconsistency can hinder knowledge transfer~\cite{huo2024c2kd,liu2023emotionkd,sun2016deepcoral}.
Frequency-based representations, including scattering transforms, wavelet pooling, and frequency-domain learning, capture complementary visual information that may not be explicit in the spatial domain
~\cite{bruna2013scattering,williams2018wavelet,xu2020frequency}. Frequency-aware distillation methods have also demonstrated that band-dependent supervision can support knowledge transfer~\cite{zhang2024freekd,pham2024frequency}. Motivated by these studies, we investigate whether band-specific alignment can complement relation-aware transfer in an intra-modal detector distillation
setting. Low-frequency responses are treated as being more associated with coarse structures, whereas high-frequency responses may be more sensitive to boundaries, fine details, and local teacher-student
discrepancies. These associations serve as design motivation and are evaluated through controlled ablations.

Although cross-attention-based feature distillation and frequency-decoupled knowledge transfer have been studied in prior work, their direct combination does not automatically define an effective detector distillation framework. Object detection requires simultaneous calibration of classification confidence, localization, and multi-scale dense feature representations. Therefore, this work does not simply introduce relation transfer and frequency-decoupled alignment as independent auxiliary objectives. Instead, they are organized as a detector-oriented refinement mechanism for large-to-small distillation. Teacher-guided alignment is evaluated not only by its fixed-budget from-scratch behavior but also by whether it forms a better student representation before subsequent detector fine-tuning. The motivation is not that prior methods universally overlook frequency information, but that a compact detector may
benefit from avoiding uniform alignment over heterogeneous transferred responses and from entering the final optimization stage with a more favorable representation.

Motivated by the above analysis, we present \textbf{Frequency-Decoupled Cross-Attention Knowledge Distillation}, referred to as \textbf{FD-CanKD}, as a detector-oriented integration framework for transferring knowledge from a high-capacity teacher to a compact student. The framework assumes that semantically corresponding multi-scale features and detector outputs can be identified between the teacher and student. It combines four components: distillation of detector-native classification and
localization outputs, multi-scale cross-attention feature relation transfer, frequency decomposition of relation-enhanced features, and band-specific feature alignment. The low-frequency branch uses MSE
alignment, while the high-frequency branch uses relaxed logarithmic alignment to reduce the influence of large local discrepancies. In detectors that employ Distribution Focal Loss, the corresponding DFL
distributions can additionally be distilled, whereas other detector families can retain their native localization supervision. This design operationalizes component-selective supervision without changing the
deployed student architecture.

Experiments on the Microsoft Common Objects in Context (COCO) dataset~\cite{lin2014coco} instantiate the proposed framework using a YOLOv12-X teacher and a YOLOv12-M student. The evaluation includes diagnostic from-scratch KD comparisons, internal component analysis, frequency-alignment ablation, scale-pair sensitivity analysis, post-distillation fine-tuning, and qualitative evaluation. The fixed 50-epoch comparisons show that FD-CanKD is competitive with representative detector KD baselines and provides the strongest \(\mathrm{mAP}_{50}\) among the compared KD variants. More importantly, the post-training fine-tuning trajectory shows that FD-CanKD provides a stronger refinement trajectory than detector-only continued fine-tuning, reaching 48.87 \(\mathrm{mAP}_{50:95}\), 65.84 \(\mathrm{mAP}_{50}\), and 53.40 \(\mathrm{mAP}_{75}\) after 20 additional epochs. These results indicate that the proposed relation-aware and frequency-aware alignment produces a student representation that can be refined more effectively, while the deployed model remains the original 19.7M-parameter YOLOv12-M student.

While the proposed principles are formulated at the level of general detector-oriented distillation, we adopt YOLOv12 as a controlled representative instantiation in this study, isolating the effect of
the proposed alignment strategy from confounding differences across detector architectures and training pipelines.

The main contributions of this paper are summarized as follows:
\begin{itemize}
    \item We present \textbf{FD-CanKD}, a detector-oriented
    large-to-small distillation framework that integrates
    detector-native output supervision, cross-attention-based relation
    transfer, and frequency-decoupled feature alignment within a
    unified training-time pipeline.

    \item We organize relation-aware and frequency-aware transfer for
    semantically corresponding multi-scale detection features.
    Instead of relying on direct point-wise matching, the compact
    student first aggregates teacher-side spatial context, after which
    the relation-enhanced features are aligned using
    component-dependent penalties.

    \item We formulate the framework independently of a particular
    detector family, provided that corresponding feature levels and
    prediction outputs can be selected and aligned. YOLOv12 is used as
    a controlled representative instantiation for empirical
    evaluation.

    \item We show that detector KD should be analyzed not only as a
    fixed-budget from-scratch training objective but also as a
    refinement starting point. Under the controlled 50-epoch
    diagnostic protocol, KD variants show modest aggregate AP
    differences, whereas KD-guided refinement clearly outperforms
    detector-only fine-tuning.

    \item We provide a three-seed reproducibility check at the
    distilled before-FT checkpoint, where FD-CanKD obtains
    48.42 \(\pm\) 0.07 \(\mathrm{mAP}_{50:95}\), compared with
    48.41 \(\pm\) 0.12 for CanKD. Continued FD-CanKD refinement
    reaches 48.87 \(\mathrm{mAP}_{50:95}\), 65.84
    \(\mathrm{mAP}_{50}\), and 53.40 \(\mathrm{mAP}_{75}\).

    \item Through controlled COCO experiments, component ablations,
    scale-pair analysis, continued fine-tuning, reproducibility checks,
    object-size-wise AP analysis, and qualitative comparisons, we show
    that FD-CanKD improves compact-detector refinement while preserving
    the original deployed YOLOv12-M architecture.
\end{itemize}

% ---------------- Related Work ----------------
\section{Related Work}

\subsection{Efficient Object Detection and YOLO Detectors}

Object detection aims to localize and classify objects under practical computation and memory constraints. Among various detector families, YOLO-based detectors have become representative efficient detection frameworks because they provide a practical balance between accuracy and efficiency. The original YOLO formulated object detection as a single-stage regression problem~\cite{redmon2016yolo}, and subsequent variants improved multi-scale prediction, feature representation, training strategies, and deployment efficiency~\cite{redmon2017yolo9000,redmon2018yolov3,bochkovskiy2020yolov4,jocher2020yolov5,li2023yolov6}. Recent YOLO variants further enhanced architectural efficiency and detection accuracy through efficient feature aggregation, anchor-free prediction, programmable gradient information, end-to-end detection, and refined latency--accuracy trade-offs~\cite{wang2023yolov7,jocher2023yolov8,wang2024yolov9,wang2024yolov10,jocher2024yolov11}. YOLOv12 extends this line by introducing an attention-centric efficient detector that improves representation quality while maintaining efficient inference~\cite{tian2025yolov12}.

Recent efficient detectors have also benefited from transformer-based designs and efficient feature processing. Denoising DETR (DN-DETR) improves DEtection TRansformer (DETR) training through query denoising, while Real-Time DETR (RT-DETR) and RT-DETRv2 improve efficient end-to-end detection through efficient encoder design and training refinements~\cite{li2022dndetr,zhao2024rtdetr,lv2024rtdetrv2}. In parallel, efficient backbone and feature aggregation designs such as Cross Stage Partial Network (CSPNet) and Efficient Layer Aggregation Network (ELAN) improve feature reuse and gradient flow~\cite{wang2020cspnet,wang2022elan}, and FlashAttention variants reduce the computational and memory overhead of attention operations~\cite{dao2022flashattention,dao2023flashattention2}. Multi-scale feature hierarchies and localization objectives are also important for accurate dense prediction. Feature Pyramid Networks provide representative multi-scale detection features~\cite{lin2017fpn}, and localization losses such as Generalized Intersection over Union (GIoU), Distance-IoU (DIoU), Intersection over Union (IoU) loss, and distribution-based regression losses improve bounding box optimization~\cite{rezatofighi2019giou,zheng2020diou,zhou2019iouloss,li2020gfl}.

Although these advances have improved detector accuracy and efficiency, compact detectors still have limited representation capacity compared with large-scale models. This limitation is critical in resource-constrained environments, where the deployed model must retain high accuracy under strict computation and memory budgets. Therefore, this paper focuses on reducing the performance gap between a large-scale YOLOv12 teacher and a compact YOLOv12 student through detector-specific knowledge distillation.

\subsection{Knowledge Distillation for Object Detection}

Knowledge distillation transfers knowledge from a high-capacity teacher model to a compact student model and has been widely used for model compression and performance improvement~\cite{bucilua2006model,hinton2015distilling}. Early studies mainly used softened output distributions or intermediate feature hints~\cite{romero2014fitnets}. Later methods improved the transfer process through collaborative learning, decoupled logit supervision, output correlation modeling, hierarchical feature review, relational constraints, Pearson correlation alignment, and selective feature imitation~\cite{zhang2018dml,zhao2022dkd,huang2022dist,chen2021review,park2019rkd,cao2022pkd,heo2019overhaul}. These studies show that effective distillation requires more than matching final outputs.

For object detection, knowledge distillation is more challenging than image classification because the student must learn both classification and localization. Detector-specific distillation has progressed along several directions. Feature-level methods transfer intermediate teacher representations to the student~\cite{li2017mimic,wang2021mimicking}, while region- and response-oriented methods emphasize foreground regions, global context, and channel-wise responses. Representative examples include Feature Knowledge Distillation (FKD)~\cite{zhang2020fkd}, Focal and Global Distillation (FGD)~\cite{yang2022fgd}, and Channel-Wise Distillation (CWD)~\cite{shu2021cwd}. Reconstruction and refinement-based methods further improve feature transfer through Masked Generative Distillation (MGD)~\cite{yang2022mgd} and knowledge diffusion~\cite{huang2023diffkd}. Other studies exploit instance-conditional supervision, distillation search, Pearson-correlation-based Knowledge Distillation (PKD), centered kernel alignment, and semantic frequency prompting, including Instance-Conditional Knowledge Distillation (ICKD)~\cite{kang2021icd}, DetKDS~\cite{li2024detkds}, Pearson-centered kernel alignment (PCKA)~\cite{zhou2024pcka}, and FreeKD~\cite{zhang2024freekd}.

Despite these advances, many detector distillation methods still treat teacher outputs or teacher features as targets to be directly matched. Prediction-level distillation transfers the final decision behavior of the teacher, but it provides limited guidance for intermediate detection features. Direct feature distillation provides dense supervision, but it often aligns teacher and student features at corresponding spatial locations without considering teacher--student spatial dependencies. When the student capacity is much smaller than the teacher capacity, such rigid feature imitation can guide the student toward responses that are not equally transferable or necessary. This limitation motivates a distillation design that preserves prediction behavior while avoiding uniform point-wise feature matching.

\subsection{Relation-Aware Feature Distillation}

Attention mechanisms are effective for modeling long-range dependencies and have become a fundamental component in modern representation learning~\cite{vaswani2017attention}. Non-local neural networks extend this idea to visual feature maps by capturing global spatial dependencies beyond local convolutional neighborhoods~\cite{wang2018nonlocal}. Relation Networks also show that object-level relationships and contextual interactions are useful for detection tasks~\cite{hu2018relation}. These studies suggest that feature distillation can benefit from modeling spatial relations rather than relying only on independent point-wise matching.

Cross-Attention-based Non-local Knowledge Distillation (CanKD) applies this idea by allowing student feature responses to attend to teacher feature responses~\cite{sun2025cankd}. Instead of forcing the student to match teacher features only at identical spatial positions, this strategy enables the student to aggregate teacher-side spatial context. This property is suitable for dense prediction tasks such as object detection because multi-scale detection features contain contextual dependencies across objects, background regions, and different spatial scales.

However, relation modeling alone does not determine how strongly each transferred feature component should be aligned. Teacher features contain heterogeneous information, including coarse structures, boundary-sensitive responses, local details, and teacher-specific variations. These components may not be equally transferable to a compact student. Therefore, we investigate a component-selective alignment strategy that controls the strength of feature imitation after spatial context has been transferred.

% ---------------- Figure 2 ----------------
\begin{figure*}[!t]
  \centering
  \includegraphics[width=\textwidth]{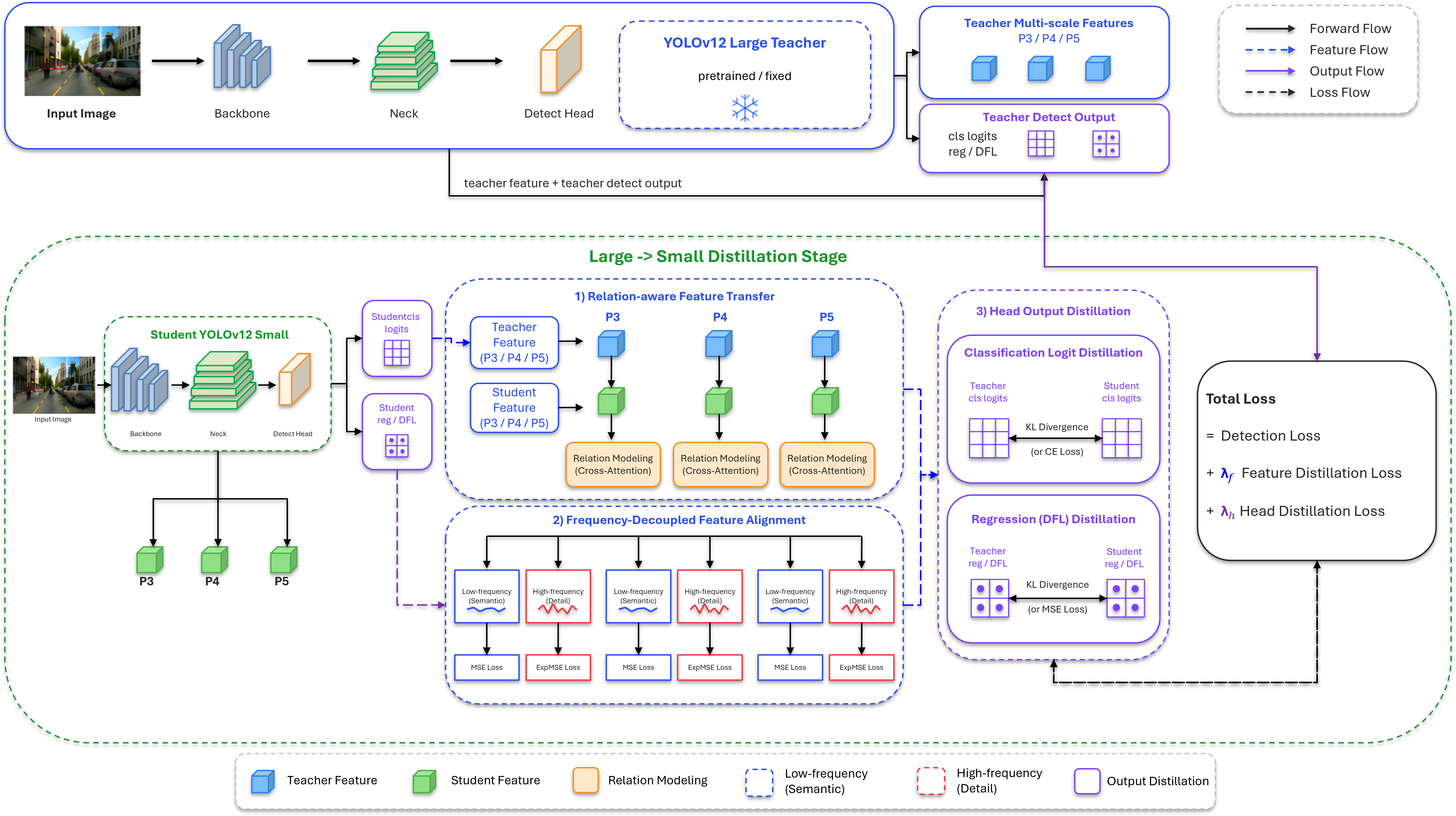}
  \caption{\textbf{Overall architecture of the FD-CanKD integration framework.} A frozen large-scale YOLOv12 teacher transfers knowledge to a compact YOLOv12 student~\cite{tian2025yolov12}. The feature-level branch performs cross-attention-based non-local distillation on multi-scale detection features~\cite{sun2025cankd}, followed by fast Fourier transform (FFT)-based decomposition into low- and high-frequency components and mean squared error (MSE) and log-compressed MSE (LogMSE) alignment~\cite{liu2025fdcmkd}. The prediction-level branch distills classification, box regression, and Distribution Focal Loss (DFL) outputs from the teacher head to the student head~\cite{li2020gfl}.}
  \label{fig:overall_fd_cankd}
\end{figure*}

\subsection{Frequency-Aware Feature Distillation}

Frequency-domain analysis provides one established way to partition feature responses into components with different spatial characteristics. Prior studies on cross-modal distillation and feature distribution alignment show that representation inconsistency can make knowledge transfer difficult~\cite{huo2024c2kd,liu2023emotionkd,sun2016deepcoral}. Frequency-based representations, including scattering transforms, wavelet pooling, and frequency-domain learning, capture complementary visual information that may not be explicit in the spatial domain~\cite{bruna2013scattering,williams2018wavelet,xu2020frequency}. Existing frequency-aware distillation methods further demonstrate the feasibility of using band-dependent supervision~\cite{zhang2024freekd,pham2024frequency}. Thus, our use of frequency decomposition builds on prior work and is introduced as a component-selective alignment mechanism.

Recent frequency-decoupled distillation studies suggest that low- and high-frequency components can benefit from different alignment strengths, particularly when representation consistency differs across components~\cite{liu2025fdcmkd}. Low-frequency responses are often associated with coarse spatial structures, whereas high-frequency responses may capture boundaries, fine local responses, and other rapidly varying details. These observations originate primarily from cross-modal settings. In this work, we treat them as a testable design motivation for intra-modal compact detector distillation rather than as a universal property of all detector features.

Building on these observations, we adopt a detector-specific band-selective design. CanKD focuses on cross-attention-based non-local feature transfer, while a recent feature-disentanglement cross-modal distillation method studies frequency-decoupled alignment for cross-modal representation discrepancy~\cite{liu2025fdcmkd}. This study organizes these complementary principles for compact YOLOv12 detector distillation. Specifically, FD-CanKD applies relation transfer to multi-scale detection features, performs band-specific alignment after relation enhancement, and jointly optimizes the feature-level terms with detection-head distillation for classification, box regression, and DFL distributions. The distinction lies in this detector-oriented ordering and coupling of relation transfer, band-specific alignment, and detection-head supervision.

% ---------------- Method ----------------
\section{Method}

\subsection{Overall Framework}

We present \textbf{Frequency-Decoupled Cross-Attention Knowledge Distillation}, referred to as \textbf{FD-CanKD}, as a detector-oriented integration framework for compact detector distillation, instantiated and controlled in this study on a YOLOv12 large-to-small setting. The objective is to improve the detection accuracy of a compact YOLOv12 student by using a high-capacity YOLOv12 teacher during training, while keeping the deployed student architecture unchanged. As shown in Figure~\ref{fig:overall_fd_cankd}, the large-scale YOLOv12 detector is used as a frozen teacher, and the compact YOLOv12 detector is optimized as the student. During inference, all distillation modules are removed, and only the compact student detector is deployed.

FD-CanKD transfers teacher knowledge through two complementary branches. The first branch performs head-output distillation to guide the student's final classification and localization behavior. The second branch performs feature-level distillation on multi-scale detection features. In this branch, student features are first projected into the teacher feature space, then enhanced by cross-attention-based relation transfer, and finally subjected to band-specific alignment using a frequency decomposition. This ordering is a key design choice: the decomposition is applied after relation-aware transfer, rather than directly on raw feature maps, so that component-selective alignment operates on context-enhanced detection representations.

Building on these prior principles, the framework focuses on their detector-oriented organization for YOLOv12 distillation. Unlike feature-only distillation, the proposed training objective couples relation transfer and component-selective alignment with detection-head distillation. This coupling is important for object detection because improved intermediate representations should also be calibrated with final classification confidence, box regression, and DFL-based localization behavior. Therefore, FD-CanKD differs from direct feature matching in two aspects. First, it transfers teacher-side spatial context instead of matching only the same pixel locations. Second, it applies different alignment rules to decomposed components after relation transfer, rather than uniformly matching all responses.

\subsection{Notation}

Let \(T_{\mathrm{L}}\) and \(S_{\mathrm{S}}\) denote the frozen large-scale YOLOv12 teacher and the trainable compact YOLOv12 student, respectively. Given an input image \(I\), both models generate detection head outputs and multi-scale detection features. The set of feature levels is denoted by \(\mathcal{P}\), corresponding to multi-scale detection features such as P3, P4, and P5. For each level \(l\in\mathcal{P}\), \(\mathbf{F}_{l}^{T}\) and \(\mathbf{F}_{l}^{S}\) denote the teacher and student feature maps.

The teacher and student head outputs contain classification logits, box regression logits, and Distribution Focal Loss (DFL)-based regression distributions. We denote the temperature-scaled class probability distributions by \(\mathbf{p}_{l}^{T}\) and \(\mathbf{p}_{l}^{S}\), the bounding box regression logits by \(\mathbf{b}_{l}^{T}\) and \(\mathbf{b}_{l}^{S}\), and the temperature-scaled DFL-based regression distributions by \(\mathbf{q}_{l}^{T}\) and \(\mathbf{q}_{l}^{S}\). The relation-enhanced student feature is denoted by \(\widehat{\mathbf{F}}_{l}^{S}\).

\subsection{Head-Output Distillation}

The head-output distillation branch provides task-level supervision for the student detector. Since object detection requires both category prediction and accurate localization, we distill three types of teacher head outputs: classification logits, bounding box regression logits, and DFL-based regression distributions. The head-output distillation loss is defined as
\begin{equation}
\begin{aligned}
\mathcal{L}_{\mathrm{head}}
=
\frac{1}{|\mathcal{P}|}
\sum_{l\in\mathcal{P}}
\Big(
&\beta_{\mathrm{cls}}
D_{\mathrm{KL}}
(\mathbf{p}_{l}^{T}\|\mathbf{p}_{l}^{S})
+
\beta_{\mathrm{reg}}
\|\mathbf{b}_{l}^{T}-\mathbf{b}_{l}^{S}\|_{1}
\\
&+
\beta_{\mathrm{dfl}}
D_{\mathrm{KL}}
(\mathbf{q}_{l}^{T}\|\mathbf{q}_{l}^{S})
\Big),
\end{aligned}
\label{eq:head_loss}
\end{equation}
where \(D_{\mathrm{KL}}(\cdot\|\cdot)\) denotes the Kullback-Leibler divergence. The coefficients \(\beta_{\mathrm{cls}}\), \(\beta_{\mathrm{reg}}\), and \(\beta_{\mathrm{dfl}}\) balance classification, regression, and DFL distillation terms, respectively. This branch guides the student to follow the teacher's final detection behavior, including class confidence, localization tendency, and distribution-based box regression.

\begin{algorithm}[t]
\caption{Simplified FD-CanKD training flow}
\label{alg:fd_cankd}
\footnotesize
\begin{algorithmic}[1]
\State \textbf{Require:} Image \(I\), label \(Y\), frozen teacher \(T_{\mathrm{L}}\), compact student \(S_{\mathrm{S}}\)
\State \textbf{Ensure:} Distilled compact student \(S_{\mathrm{S}}\)
\State Extract teacher outputs and features: \((\mathcal{O}^{T}, \{\mathbf{F}^{T}\}) \leftarrow T_{\mathrm{L}}(I)\)
\State Extract student outputs and features: \((\mathcal{O}^{S}, \{\mathbf{F}^{S}\}) \leftarrow S_{\mathrm{S}}(I)\)
\State Compute detection loss: \(\mathcal{L}_{\mathrm{det}} \leftarrow \mathrm{YOLOLoss}(\mathcal{O}^{S},Y)\)
\State Compute head-output distillation loss: \(\mathcal{L}_{\mathrm{head}} \leftarrow \mathrm{HeadKD}(\mathcal{O}^{T},\mathcal{O}^{S})\)
\State Apply relation-aware transfer: \(\{\widehat{\mathbf{F}}^{S}\} \leftarrow \operatorname{RelAttn}(\mathcal{A}(\{\mathbf{F}^{S}\}),\{\mathbf{F}^{T}\})\)
\State Decompose \(\{\mathbf{F}^{T}\}\) and \(\{\widehat{\mathbf{F}}^{S}\}\) into low- and high-frequency components
\State Compute frequency-decoupled loss: \(\mathcal{L}_{\mathrm{fd}} \leftarrow \mathrm{MSE}_{\mathrm{low}}+\mathrm{LogMSE}_{\mathrm{high}}\)
\State Update \(S_{\mathrm{S}}\) using \(\mathcal{L}_{\mathrm{total}}\)
\State Remove distillation modules and deploy \(S_{\mathrm{S}}\)
\end{algorithmic}
\end{algorithm}

\subsection{Multi-Scale Relation-Aware Feature Transfer}

Direct feature matching aligns teacher and student features at corresponding spatial locations. However, detection features contain contextual dependencies across object regions, surrounding background, and different object scales. To transfer such contextual information, we apply cross-attention-based relation transfer to multi-scale YOLOv12 detection features.

For each feature level \(l\), the student feature is first projected to the teacher channel dimension using a \(1 \times 1\) adapter \(\mathcal{A}_{l}(\cdot)\). The adapted student feature attends to the teacher feature and produces the relation-enhanced student feature:
\begin{equation}
\begin{aligned}
&\widetilde{\mathbf{F}}_{l}^{S}
=
\mathcal{A}_{l}(\mathbf{F}_{l}^{S}),
\\
&\widehat{\mathbf{f}}_{l,i}^{S}
=
\frac{1}{N_l}
\sum_{j=1}^{N_l}
\xi(\widetilde{\mathbf{f}}_{l,i}^{S},\mathbf{f}_{l,j}^{T})
g(\mathbf{f}_{l,j}^{T}),
\\
&\xi(\widetilde{\mathbf{f}}_{l,i}^{S},\mathbf{f}_{l,j}^{T})
=
\theta(\widetilde{\mathbf{f}}_{l,i}^{S})^{\top}
\phi(\mathbf{f}_{l,j}^{T}),
\end{aligned}
\label{eq:relation_transfer}
\end{equation}
where \(N_l=H_lW_l\), and \(\widetilde{\mathbf{f}}_{l,i}^{S}\) and \(\mathbf{f}_{l,j}^{T}\) are feature vectors at spatial positions \(i\) and \(j\), respectively. The functions \(\theta(\cdot)\), \(\phi(\cdot)\), and \(g(\cdot)\) are implemented using \(1 \times 1\) convolutional projections. Following CanKD~\cite{sun2025cankd}, we use an unnormalized dot-product affinity with the factor \(1/N_l\), rather than softmax-normalized attention. This formulation preserves direct relative response differences across teacher positions and avoids introducing an additional probability normalization into the cross-attention non-local operation. The output vectors \(\{\widehat{\mathbf{f}}_{l,i}^{S}\}_{i=1}^{N_l}\) are reshaped into \(\widehat{\mathbf{F}}_{l}^{S}\). Through this operation, the student aggregates teacher-side non-local context rather than imitating only the corresponding spatial location.

\subsection{Frequency-Decoupled Feature Alignment}

Although relation-aware transfer provides spatially enriched feature guidance, the transferred representation still contains heterogeneous responses. Motivated by prior frequency-aware distillation studies, we use frequency decomposition as a practical partitioning mechanism for applying component-dependent supervision. Low-frequency responses are often associated with coarse spatial structures, whereas high-frequency responses may be more sensitive to boundaries, local activations, and teacher--student discrepancies. These associations motivate, but do not by themselves establish, the use of different alignment rules.

Accordingly, we decompose the teacher feature \(\mathbf{F}_{l}^{T}\) and the relation-enhanced student feature \(\widehat{\mathbf{F}}_{l}^{S}\) into low- and high-frequency components. Before the fast Fourier transform (FFT), we remove the channel-wise spatial mean as the direct-current (DC) component to reduce domination by the global feature magnitude. The frequency decomposition is defined as
\begin{equation}
\begin{aligned}
\mathbf{U}_{l}^{m}
&=
\operatorname{fftshift}
\left(
\mathcal{F}_{2D}
\left(
\mathcal{D}(\mathbf{F}_{l}^{m})
\right)
\right),
\\
\mathbf{F}_{l,r}^{m}
&=
\operatorname{Re}
\left[
\mathcal{F}_{2D}^{-1}
\left(
\operatorname{ifftshift}
\left(
\mathbf{U}_{l}^{m}
\odot
\mathbf{M}_{r}
\right)
\right)
\right],
\end{aligned}
\label{eq:frequency_decomposition}
\end{equation}
where \(m\in\{T,\widehat{S}\}\), \(r\in\{\mathrm{low},\mathrm{high}\}\), \(\mathcal{D}(\cdot)\) denotes DC filtering, \(\mathcal{F}_{2D}(\cdot)\) and \(\mathcal{F}_{2D}^{-1}(\cdot)\) denote the 2D FFT and inverse 2D FFT, respectively, and \(\mathbf{M}_{r}\) is the corresponding frequency mask. The \(\operatorname{fftshift}(\cdot)\) operation moves the zero-frequency component to the center of the spectrum so that the radial masks in Equations~\eqref{eq:radial_distance} and~\eqref{eq:frequency_mask} are centered at \((H_l/2, W_l/2)\). Before the inverse transform, \(\operatorname{ifftshift}(\cdot)\) restores the original frequency ordering.

To define the low- and high-frequency masks compactly, we use the normalized radial frequency distance:
\begin{equation}
d_l(u, v)
=
\left\|
\left(
\frac{u-H_l/2}{H_l},
\frac{v-W_l/2}{W_l}
\right)
\right\|_2.
\label{eq:radial_distance}
\end{equation}
The low- and high-frequency masks are then defined as
\begin{equation}
\mathbf{M}_{\mathrm{low}}(u, v)
=
\mathbb{I}
\left[
d_l(u, v)\leq\tau
\right],
\quad
\mathbf{M}_{\mathrm{high}}
=
1-\mathbf{M}_{\mathrm{low}},
\label{eq:frequency_mask}
\end{equation}
where \((u, v)\) denotes the frequency coordinate, \(\mathbb{I}[\cdot]\) is the indicator function, and \(\tau\) is the cutoff ratio. In our implementation, \(\tau=0.25\) is used unless otherwise specified.

% ==================== TABLE 1 START: Controlled comparison with representative detector KD methods ====================
\begin{table*}[!t]
\centering
\footnotesize
\setlength{\tabcolsep}{4pt}
\renewcommand{\arraystretch}{1.05}
\caption{Controlled comparison with the locally reproduced 50-epoch from-scratch YOLOv12-M student reference and representative detector KD methods under the same YOLOv12-X to YOLOv12-M local training protocol. External feature-distillation methods are combined with the same head-output distillation branch for a fair controlled comparison.}
\label{tab:external_kd_comparison}
\begin{tabularx}{\textwidth}{@{}lXcccc@{}}
\toprule
\textbf{Method} & \textbf{Feature KD Target} & \textbf{Head KD} & \(\mathbf{mAP}_{50:95}\) & \(\mathbf{mAP}_{50}\) & \(\mathbf{mAP}_{75}\) \\
\midrule
YOLOv12-M Student (50-epoch FS) & -- & No & 45.6 & 62.0 & 50.8 \\
CWD + Head KD~\cite{shu2021cwd} & Channel-wise distribution & Yes & 46.3 & 62.4 & 50.5 \\
MGD + Head KD~\cite{yang2022mgd} & Masked generative feature & Yes & 46.3 & 62.3 & 50.4 \\
PKD + Head KD~\cite{cao2022pkd} & Pearson correlation & Yes & 46.2 & 62.2 & 50.5 \\
FGD + Head KD~\cite{yang2022fgd} & Focal/global feature & Yes & 46.3 & 62.2 & 50.4 \\
FreeKD + Head KD~\cite{zhang2024freekd} & Frequency-domain feature & Yes & 46.5 & 62.7 & 50.7 \\
\textbf{FD-CanKD} & \textbf{Relation + frequency} & \textbf{Yes} & \textbf{46.5} & \textbf{62.8} & \textbf{50.8} \\
\bottomrule
\end{tabularx}
\end{table*}
% ==================== TABLE 1 END ====================

% ==================== TABLE 2 START: Internal component comparison ====================
\begin{table*}[!t]
\centering
\footnotesize
\setlength{\tabcolsep}{4pt}
\renewcommand{\arraystretch}{1.05}
\caption{Selected internal component comparison under the controlled 50-epoch from-scratch YOLOv12-X to YOLOv12-M local training protocol. The table separates relation-aware feature transfer, head-output distillation, and frequency-aware alignment.}
\label{tab:yolov12_overall_comparison}
\begin{tabular*}{0.82\textwidth}{@{\extracolsep{\fill}}lcccccc@{}}
\toprule
\textbf{Method} & \textbf{Rel.} & \textbf{Head} & \textbf{Freq.} & \(\mathbf{mAP}_{50:95}\) & \(\mathbf{mAP}_{50}\) & \(\mathbf{mAP}_{75}\) \\
\midrule
YOLOv12-M Student (50-epoch FS) & -- & -- & -- & 45.6 & 62.0 & 50.8 \\
CanKD~\cite{sun2025cankd} & Yes & No & No & 46.5 & 62.4 & 50.5 \\
CanKD + Head KD & Yes & Yes & No & 46.3 & 62.4 & 50.5 \\
\textbf{FD-CanKD} & \textbf{Yes} & \textbf{Yes} & \textbf{Yes} & \textbf{46.5} & \textbf{62.8} & \textbf{50.8} \\
\bottomrule
\end{tabular*}

\vspace{2pt}
\begin{minipage}{0.82\textwidth}
\footnotesize\textit{Note.} Rel., relation-aware feature transfer. Head, head-output distillation. Freq., frequency-aware alignment.
\end{minipage}
\end{table*}
% ==================== TABLE 2 END ====================

The low-frequency branch is aligned with MSE to preserve broad structural correspondence. The high-frequency branch is aligned with a logarithmic mapping that limits the influence of large local discrepancies. Let \(\bar{\mathbf{F}}\) denote the channel-wise normalized feature map, and let \(\rho(x)=\operatorname{sign}(x)\log(1+|x|)\). The frequency-decoupled feature distillation loss is defined as
\begin{equation}
\begin{aligned}
\mathcal{L}_{\mathrm{fd}}
=
\frac{1}{|\mathcal{P}|}
\sum_{l\in\mathcal{P}}
\Big(
&\gamma_{\mathrm{low}}
\|
\bar{\mathbf{F}}_{l,\mathrm{low}}^{T}
-
\bar{\mathbf{F}}_{l,\mathrm{low}}^{\widehat{S}}
\|_{2}^{2}
\\
&+
\gamma_{\mathrm{high}}
\|
\rho(
\bar{\mathbf{F}}_{l,\mathrm{high}}^{T}
-
\bar{\mathbf{F}}_{l,\mathrm{high}}^{\widehat{S}}
)
\|_{2}^{2}
\Big),
\end{aligned}
\label{eq:fd_loss}
\end{equation}
where \(\gamma_{\mathrm{low}}\) and \(\gamma_{\mathrm{high}}\) balance the low- and high-frequency alignment terms. This formulation corresponds to MSE alignment for low-frequency components and LogMSE alignment for high-frequency components. The resulting objective implements different penalties for broad structural correspondence and rapidly varying local discrepancies.

\subsection{Overall Objective and Training Flow}

The final objective combines the original YOLOv12 detection loss, the head-output distillation loss, and the frequency-decoupled feature distillation loss. Following the YOLOv12 training objective~\cite{tian2025yolov12}, the original detection loss consists of box regression, classification, and DFL terms. The total training objective is defined as
\begin{equation}
\begin{aligned}
\mathcal{L}_{\mathrm{det}}
&=
\alpha_{\mathrm{box}}\mathcal{L}_{\mathrm{box}}
+
\alpha_{\mathrm{cls}}\mathcal{L}_{\mathrm{cls}}
+
\alpha_{\mathrm{dfl}}\mathcal{L}_{\mathrm{dfl}},
\\
\mathcal{L}_{\mathrm{total}}
&=
\mathcal{L}_{\mathrm{det}}
+
\lambda_{\mathrm{head}}\mathcal{L}_{\mathrm{head}}
+
\lambda_{\mathrm{fd}}\mathcal{L}_{\mathrm{fd}}.
\end{aligned}
\label{eq:total_loss}
\end{equation}
Here, \(\lambda_{\mathrm{head}}\) and \(\lambda_{\mathrm{fd}}\) control the contribution of head-output and feature-level distillation losses, respectively.

Algorithm~\ref{alg:fd_cankd} summarizes the training flow corresponding to Figure~\ref{fig:overall_fd_cankd}. The teacher provides head-output and feature-level supervision only during training. After distillation, all training-time modules are removed, and only the compact YOLOv12 student is deployed. Therefore, FD-CanKD improves the compact detector through training-time supervision without changing the deployed YOLOv12-M architecture, which is important for resource-constrained intelligent vision systems.

% ---------------- Experiments ----------------

\section{Experiments}

\subsection{Experimental Setup}
\label{sec:experimental_setup}

We evaluate the FD-CanKD framework on the Microsoft Common Objects in Context (COCO) dataset~\cite{lin2014coco}. The main setting uses YOLOv12-X as the teacher model and YOLOv12-M as the student model. During distillation, the teacher network is frozen, and only the student network and training-time distillation modules are optimized. Unless otherwise stated, all experiments are conducted with an input resolution of 640, a batch size of 16, and 50 training epochs. The main evaluation metrics are mean average precision over IoU thresholds from 0.50 to 0.95 (\(\mathrm{mAP}_{50:95}\)), \(\mathrm{mAP}_{50}\), and \(\mathrm{mAP}_{75}\) on the COCO validation set.

% ==================== TABLE 3 START: Frequency-decoupled alignment ablation ====================
\begin{table*}[!t]
\centering
\footnotesize
\setlength{\tabcolsep}{4pt}
\renewcommand{\arraystretch}{1.05}
\caption{Frequency-decoupled alignment ablation after relation-aware feature transfer under the controlled 50-epoch from-scratch teacher--student setting.}
\label{tab:frequency_ablation}
\begin{tabular*}{0.90\textwidth}{@{\extracolsep{\fill}}lccccccc@{}}
\toprule
\textbf{Method} & \textbf{Low} & \textbf{High} & \textbf{Low Loss} & \textbf{High Loss} & \(\mathbf{mAP}_{50:95}\) & \(\mathbf{mAP}_{50}\) & \(\mathbf{mAP}_{75}\) \\
\midrule
YOLOv12-M Student (50-epoch FS) & -- & -- & -- & -- & 45.6 & 62.0 & 50.8 \\
Rel. KD + Head & -- & -- & -- & -- & 46.3 & 62.4 & 50.5 \\
All-MSE & Yes & Yes & MSE & MSE & 46.5 & 62.4 & 50.6 \\
Low-only & Yes & No & MSE & -- & 46.5 & 62.6 & 50.7 \\
High-only & No & Yes & -- & LogMSE & 46.4 & 62.5 & 50.3 \\
\textbf{Low + High-log} & \textbf{Yes} & \textbf{Yes} & \textbf{MSE} & \textbf{LogMSE} & \textbf{46.5} & \textbf{62.4} & \textbf{50.8} \\
\bottomrule
\end{tabular*}

\vspace{2pt}
\begin{minipage}{0.90\textwidth}
\footnotesize\textit{Note.} MSE, mean squared error. LogMSE, log-compressed mean squared error.
\end{minipage}
\end{table*}
% ==================== TABLE 3 END ====================

% ==================== TABLE 4 START: Scale-pair sensitivity analysis ====================
\begin{table*}[!t]
\centering
\footnotesize
\setlength{\tabcolsep}{4pt}
\renewcommand{\arraystretch}{1.05}
\caption{Scale-pair sensitivity analysis across different YOLOv12 teacher--student pairs under the controlled 50-epoch from-scratch local protocol.}
\label{tab:scale_pair_generalization}
\begin{tabular*}{0.78\textwidth}{@{\extracolsep{\fill}}lccc@{}}
\toprule
\textbf{Scale Pair} & \textbf{Teacher \(\mathbf{mAP}_{50:95}\)} & \textbf{Student Base \(\mathbf{mAP}_{50:95}\)} & \textbf{FD-CanKD \(\mathbf{mAP}_{50:95}\)} \\
\midrule
YOLOv12-L \(\rightarrow\) YOLOv12-M & 47.0 & 45.6 & 46.4 \\
YOLOv12-X \(\rightarrow\) YOLOv12-M & 49.5 & 45.6 & 46.5 \\
YOLOv12-X \(\rightarrow\) YOLOv12-S & 49.5 & 40.9 & 41.1 \\
\bottomrule
\end{tabular*}

\vspace{2pt}
\begin{minipage}{0.78\textwidth}
\footnotesize\textit{Note.} All AP values are obtained from the controlled local protocol described in Section~\ref{sec:experimental_setup}.
\end{minipage}
\end{table*}
% ==================== TABLE 4 END ====================

The official YOLOv12 study reports COCO results obtained using a substantially longer 600-epoch training schedule~\cite{tian2025yolov12}. In contrast, the objective of the present experiments is not to reproduce the official YOLOv12 leaderboard, but to verify the applicability and component interactions of FD-CanKD under a controlled local environment. Accordingly, the YOLOv12-X, YOLOv12-M, and YOLOv12-S reference models, representative detector KD baselines, and internal distillation variants are evaluated using the same local codebase, input resolution, batch size, 50-epoch training duration, data-processing pipeline, and validation rule. The absolute AP values reported in this paper should therefore be interpreted as internally controlled local results rather than direct reproductions of the official YOLOv12 values.

For fair comparison, the representative external KD comparisons and internal component ablations follow the same from-scratch local protocol. In this setting, all student-side models are trained from scratch under the same initialization policy, training schedule, data-processing pipeline, and validation rule. The 50-epoch YOLOv12-M student is used as a fixed from-scratch reference for Tables~\ref{tab:external_kd_comparison}--\ref{tab:scale_pair_generalization}. It should not be interpreted as a fully converged standalone detector baseline. Table~\ref{tab:finetune_warmstart} separately evaluates continued detector optimization under a post-training analysis with its own locally evaluated student and teacher references. This separation prevents the diagnostic from-scratch KD comparisons in Tables~\ref{tab:external_kd_comparison}--\ref{tab:scale_pair_generalization} from being conflated with the warm-start distillation and continued fine-tuning trajectories in Table~\ref{tab:finetune_warmstart}. We additionally evaluate post-distillation fine-tuning and a warm-start distillation protocol to examine whether teacher-guided feature alignment becomes more effective after the compact detector has already learned a stable detection representation. In the warm-start FD-CanKD setting, the student is initialized from locally trained YOLOv12-M student weights and then optimized with the proposed distillation objective under the same local training pipeline.

The proposed method is implemented on top of a YOLOv12/Ultralytics-style training pipeline. For reproducibility, the main implementation hyperparameters are fixed across the controlled comparisons rather than tuned separately for each baseline or ablation. The values are selected as stable local settings for the 50-epoch controlled protocol and are kept unchanged throughout the experiments. The frequency cutoff is set to \(\tau=0.25\), which defines a compact centered low-frequency region while leaving the remaining spectrum to the high-frequency branch. For head-output distillation, the classification, box-regression, and DFL weights are set to \(\beta_{\mathrm{cls}}=1.0\), \(\beta_{\mathrm{reg}}=0.0\), and \(\beta_{\mathrm{dfl}}=1.0\), respectively. Thus, the main implementation uses classification and DFL distribution distillation and does not apply a separate L1 box-regression KD term, avoiding an additional box-logit constraint beyond the original YOLO detection loss. For frequency-decoupled feature alignment, the low- and high-frequency weights are set to \(\gamma_{\mathrm{low}}=1.0\) and \(\gamma_{\mathrm{high}}=0.5\), so that broad structural correspondence receives the full MSE penalty while detail-sensitive high-frequency discrepancies are softened. The total loss weights are set to \(\lambda_{\mathrm{head}}=0.2\) and \(\lambda_{\mathrm{fd}}=0.1\), keeping distillation as auxiliary supervision relative to the primary detection objective. The multi-scale feature weights for P3/P4/P5 are \(0.25/1.0/1.0\), which reduces the influence of the highest-resolution P3 feature and emphasizes the more semantically stable P4/P5 detection features. The head-KD temperature is \(T=2.0\), and the teacher confidence threshold is 0.1 to retain broad teacher guidance while filtering extremely low-confidence predictions. The original YOLO detection loss weights are \(\alpha_{\mathrm{box}}=7.5\), \(\alpha_{\mathrm{cls}}=0.5\), and \(\alpha_{\mathrm{dfl}}=1.5\). A 5-epoch feature warm-up and a 3-epoch head warm-up are used, with late-stage decay factors applied to stabilize distillation after the student has formed initial detection responses.

In addition to the original YOLO detection loss, FD-CanKD applies head-output distillation and feature-level distillation. The feature-level branch operates on multi-scale detection features and combines cross-attention-based relation transfer with frequency-aware alignment. After training, all distillation modules are removed, and only the compact student detector is used for inference.

% ==================== TABLE 5 START: Post-distillation fine-tuning trajectory and reference models ====================
\begin{table*}[!t]
\centering
\footnotesize
\setlength{\tabcolsep}{3.2pt}
\renewcommand{\arraystretch}{1.05}
\caption{Post-distillation continued fine-tuning analysis under the controlled local protocol. The table compares detector-only continued fine-tuning with warm-start CanKD and FD-CanKD trajectories, together with locally evaluated student and teacher references. The distilled before-FT rows report the mean and sample standard deviation over three random seeds.}
\label{tab:finetune_warmstart}
\begin{tabular*}{0.96\textwidth}{@{\extracolsep{\fill}}lllccc@{}}
\toprule
\textbf{Method} & \textbf{Init} & \textbf{Stage / Epochs} & \(\mathbf{mAP}_{50:95}\) & \(\mathbf{mAP}_{50}\) & \(\mathbf{mAP}_{75}\) \\
\midrule
\multirow{4}{*}{YOLOv12-M FT}
& \multirow{4}{*}{--}
& +5 FT  & 46.66 & 63.15 & 50.91 \\
& & +10 FT & 46.71 & 63.25 & 50.81 \\
& & +20 FT & 46.90 & 63.50 & 51.01 \\
& & +30 FT & \textbf{47.02} & \textbf{63.68} & \textbf{51.20} \\
\midrule
\multirow{5}{*}{CanKD}
& \multirow{5}{*}{WS}
& Distilled, before FT & 48.41 \(\pm\) 0.12 & 65.09 \(\pm\) 0.13 & 52.79 \(\pm\) 0.18 \\
& & +5 FT  & 48.50 & 65.23 & 52.76 \\
& & +10 FT & 48.58 & 65.38 & 52.86 \\
& & +20 FT & 48.75 & 65.56 & 52.94 \\
& & +30 FT & \textbf{48.80} & \textbf{65.65} & \textbf{53.19} \\
\midrule
\multirow{5}{*}{\textbf{FD-CanKD}}
& \multirow{5}{*}{WS}
& Distilled, before FT & 48.42 \(\pm\) 0.07 & 65.00 \(\pm\) 0.13 & 52.85 \(\pm\) 0.07 \\
& & +5 FT  & 48.72 & 65.52 & 53.10 \\
& & +10 FT & 48.74 & 65.65 & 53.13 \\
& & +20 FT & \textbf{48.87} & \textbf{65.84} & \textbf{53.40} \\
& & +30 FT & 48.84 & 65.82 & 53.35 \\
\midrule
YOLOv12-M Student & FS & Local student (50-ep FS ref.) & 45.63 & 62.01 & 50.81 \\
YOLOv12-X Teacher & -- & Local teacher (ref.) & 49.58 & 66.29 & 54.04 \\
\bottomrule
\end{tabular*}

\vspace{2pt}
\begin{minipage}{0.96\textwidth}
\footnotesize\textit{Note.} Init denotes the initialization used for the corresponding training or distillation stage. FS denotes from-scratch training with random initialization for 50 epochs, whereas WS denotes warm-start initialization from the locally trained 50-epoch YOLOv12-M student before the 50-epoch distillation stage. The locally evaluated student and teacher rows are references specific to this post-training analysis. The 45.63 local student reference is reused consistently in Table~\ref{tab:object_size_analysis}. The 45.6 diagnostic from-scratch reference in Tables~\ref{tab:external_kd_comparison}--\ref{tab:scale_pair_generalization} belongs to the fixed 50-epoch diagnostic comparison protocol and should not be used interchangeably for cross-table gain calculations. For the CanKD and FD-CanKD ``Distilled, before FT'' rows, values are reported as mean \(\pm\) sample standard deviation over three random seeds. FT denotes continued detector fine-tuning without KD loss, and the +5/+10/+20/+30 FT rows denote additional detector fine-tuning epochs from the corresponding starting point. The YOLOv12-X teacher row is a locally evaluated reference and is not a deployed student result.
\end{minipage}
\end{table*}
% ==================== TABLE 5 END ====================

% ==================== TABLE 6 START: Object-size-wise analysis ====================
\begin{table*}[!t]
\centering
\footnotesize
\setlength{\tabcolsep}{3.8pt}
\renewcommand{\arraystretch}{1.05}
\caption{Object-size-wise COCO AP for the local YOLOv12-M student reference and the +20- and +30-epoch continued fine-tuning checkpoints.}
\label{tab:object_size_analysis}
\begin{tabular*}{0.96\textwidth}{@{\extracolsep{\fill}}lllcccc@{}}
\toprule
\textbf{Method} & \textbf{Stage} & \textbf{Setting} & \(\mathbf{mAP}_{50:95}\) & \(\mathbf{AP}_{S}\) & \(\mathbf{AP}_{M}\) & \(\mathbf{AP}_{L}\) \\
\midrule
YOLOv12-M Student (ref.) & Reference & Local student & 45.63 & 28.92 & 51.94 & 62.35 \\
\midrule
CanKD & +20 FT & Relation-aware & 48.75 & 29.19 & \textbf{53.76} & \textbf{66.09} \\
\textbf{FD-CanKD} & \textbf{+20 FT} & \textbf{Relation + frequency} & \textbf{48.87} & \textbf{30.15} & 53.54 & 65.41 \\
\midrule
CanKD & +30 FT & Relation-aware & 48.80 & 29.44 & \textbf{53.87} & \textbf{66.23} \\
\textbf{FD-CanKD} & \textbf{+30 FT} & \textbf{Relation + frequency} & \textbf{48.84} & \textbf{30.02} & 53.43 & 65.43 \\
\bottomrule
\end{tabular*}

\vspace{2pt}
\begin{minipage}{0.96\textwidth}
\footnotesize\textit{Note.} FT denotes continued detector fine-tuning. The YOLOv12-M Student (ref.) row reproduces the same 45.63 local student reference used in Table~\ref{tab:finetune_warmstart} and serves only as a common anchor. It is not a continued fine-tuning checkpoint. The +20 and +30 FT rows are matched-stage comparisons between CanKD and FD-CanKD from their corresponding warm-start distillation trajectories. \(\mathrm{AP}_{S}\), \(\mathrm{AP}_{M}\), and \(\mathrm{AP}_{L}\) follow the COCO area definitions for small, medium, and large objects, respectively~\cite{lin2014coco}. Two neighboring checkpoints are reported to reduce dependence on a single selected stage. All entries are evaluation results from the corresponding saved models. No additional training is performed for this analysis.
\end{minipage}
\end{table*}
% ==================== TABLE 6 END ====================

\subsection{Controlled Comparison and Ablation Protocol}

We evaluate FD-CanKD through representative external comparisons and controlled ablations under the same local YOLOv12-X to YOLOv12-M protocol. Because detector distillation results are sensitive to codebase, schedule, initialization, and evaluation details, the teacher, student, representative external KD baselines, and internal ablation variants are all reproduced within the same local environment. This design allows the effect of the proposed relation-plus-frequency alignment to be compared against representative detector distillation targets without conflating it with differences in training duration, implementation, or validation protocol.

The quantitative evaluation is organized into six parts. Tables~\ref{tab:external_kd_comparison}--\ref{tab:scale_pair_generalization} are interpreted as diagnostic comparisons under a fixed 50-epoch from-scratch local protocol. They examine representative detector KD targets, selected internal components, the frequency-decoupled alignment rule, and teacher--student scale sensitivity without using extra detector-only optimization. The post-training behavior is evaluated separately in Table~\ref{tab:finetune_warmstart}, where detector-only fine-tuning and KD-guided fine-tuning are compared from their corresponding local starting points. Random-seed variability at the distilled before-FT checkpoint is reported directly in Table~\ref{tab:finetune_warmstart} using the sample standard deviation over three random seeds and is discussed as a robustness check rather than as a formal significance test. Finally, Table~\ref{tab:object_size_analysis} reports the same local YOLOv12-M student reference used in Table~\ref{tab:finetune_warmstart}, together with a COCO object-size-wise AP breakdown at the matched +20- and +30-epoch continued fine-tuning checkpoints. The reference row serves only as a common anchor, whereas the size-specific method comparison is made between CanKD and FD-CanKD at the same continued fine-tuning stage. This organization identifies whether the observed FD-CanKD difference is concentrated in a particular object-size regime and whether the trend persists across neighboring refinement stages.

% ---------------- Figure 3 ----------------
\begin{figure*}[!t]
  \centering
  \includegraphics[width=\textwidth]{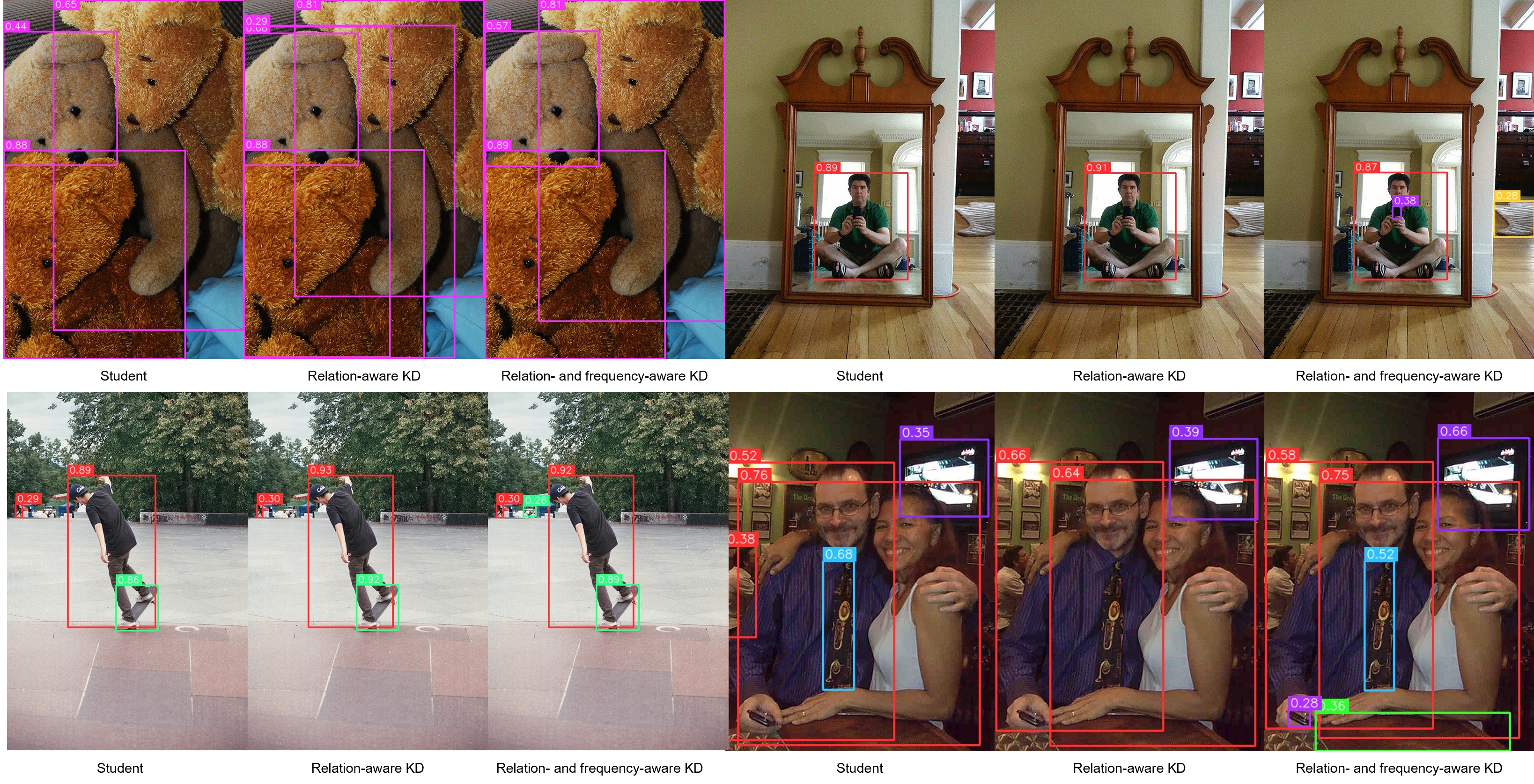}
  \caption{\textbf{Qualitative comparison on cluttered and occluded scenes.} From left to right in each group, the results correspond to the YOLOv12-M student, the relation-aware KD variant, and FD-CanKD. The baseline student often shows lower confidence or unstable predictions in cluttered regions, while relation-aware KD improves several object responses. FD-CanKD combines relation-aware feature transfer with band-specific alignment and shows more stable detections, including fine-grained local object responses, in the illustrated examples.}
  \label{fig:qualitative_clutter}
\end{figure*}

\subsection{Overall Performance Comparison}

Table~\ref{tab:external_kd_comparison} summarizes the from-scratch diagnostic comparison among the locally reproduced 50-epoch YOLOv12-M student reference, representative detector distillation methods, and FD-CanKD under the same local YOLOv12-X to YOLOv12-M protocol.

Under this fixed 50-epoch setting, the locally reproduced YOLOv12-M student reference obtains 45.6 \(\mathrm{mAP}_{50:95}\), 62.0 \(\mathrm{mAP}_{50}\), and 50.8 \(\mathrm{mAP}_{75}\). This value is used as a controlled from-scratch reference for comparing KD variants, not as a fully optimized detector-only baseline. CWD, MGD, PKD, FGD, and FreeKD combined with head-output distillation obtain \(\mathrm{mAP}_{50:95}\) values between 46.2 and 46.5. FD-CanKD obtains 46.5 \(\mathrm{mAP}_{50:95}\), while reaching 62.8 \(\mathrm{mAP}_{50}\) and 50.8 \(\mathrm{mAP}_{75}\). Although the aggregate AP difference among KD variants is modest, FD-CanKD gives the strongest \(\mathrm{mAP}_{50}\) while preserving the student-level \(\mathrm{mAP}_{75}\). Thus, Table~\ref{tab:external_kd_comparison} is used to analyze the controlled distillation behavior under the same training budget, while the effect of additional detector-only optimization is evaluated separately in Table~\ref{tab:finetune_warmstart}.

\subsection{Internal Component Ablation}

Table~\ref{tab:yolov12_overall_comparison} analyzes the selected internal components under the from-scratch setting.

The locally reproduced 50-epoch YOLOv12-M student reference obtains 45.6 \(\mathrm{mAP}_{50:95}\). CanKD alone reaches 46.5, and adding head-output distillation reaches 46.3. The full from-scratch FD-CanKD also reaches 46.5 while producing the strongest \(\mathrm{mAP}_{50}\) among the internal variants and preserving \(\mathrm{mAP}_{75}\) relative to the student. These results show that relation transfer and frequency-aware alignment provide useful guidance under the controlled protocol, while the small numerical differences among internal variants suggest that aggregate AP alone does not fully explain the qualitative behavior of the detectors.

\subsection{Frequency-Decoupled Alignment Ablation}

Table~\ref{tab:frequency_ablation} analyzes alternative alignment rules for the decomposed feature components under the from-scratch setting.

The locally reproduced 50-epoch YOLOv12-M student reference obtains 45.6 \(\mathrm{mAP}_{50:95}\), 62.0 \(\mathrm{mAP}_{50}\), and 50.8 \(\mathrm{mAP}_{75}\). The Rel. KD + Head baseline obtains 46.3 \(\mathrm{mAP}_{50:95}\), 62.4 \(\mathrm{mAP}_{50}\), and 50.5 \(\mathrm{mAP}_{75}\). The All-MSE and Low-only settings both reach 46.5 \(\mathrm{mAP}_{50:95}\), while the High-only setting reaches 46.4. The final Low + High-log setting reaches 46.5 \(\mathrm{mAP}_{50:95}\), 62.4 \(\mathrm{mAP}_{50}\), and 50.8 \(\mathrm{mAP}_{75}\). Although the aggregate AP differences are small, the asymmetric objective preserves the student-level \(\mathrm{mAP}_{75}\) while improving \(\mathrm{mAP}_{50:95}\) over the student reference. This supports the design choice that low-frequency responses, which tend to encode broader object-level structure and semantic context, can be directly aligned with MSE, whereas high-frequency responses, which contain boundary and local-detail variations, benefit from a softened log-compressed penalty.

% ---------------- Figure 4 ----------------
\begin{figure*}[!t]
  \centering
  \includegraphics[width=\textwidth]{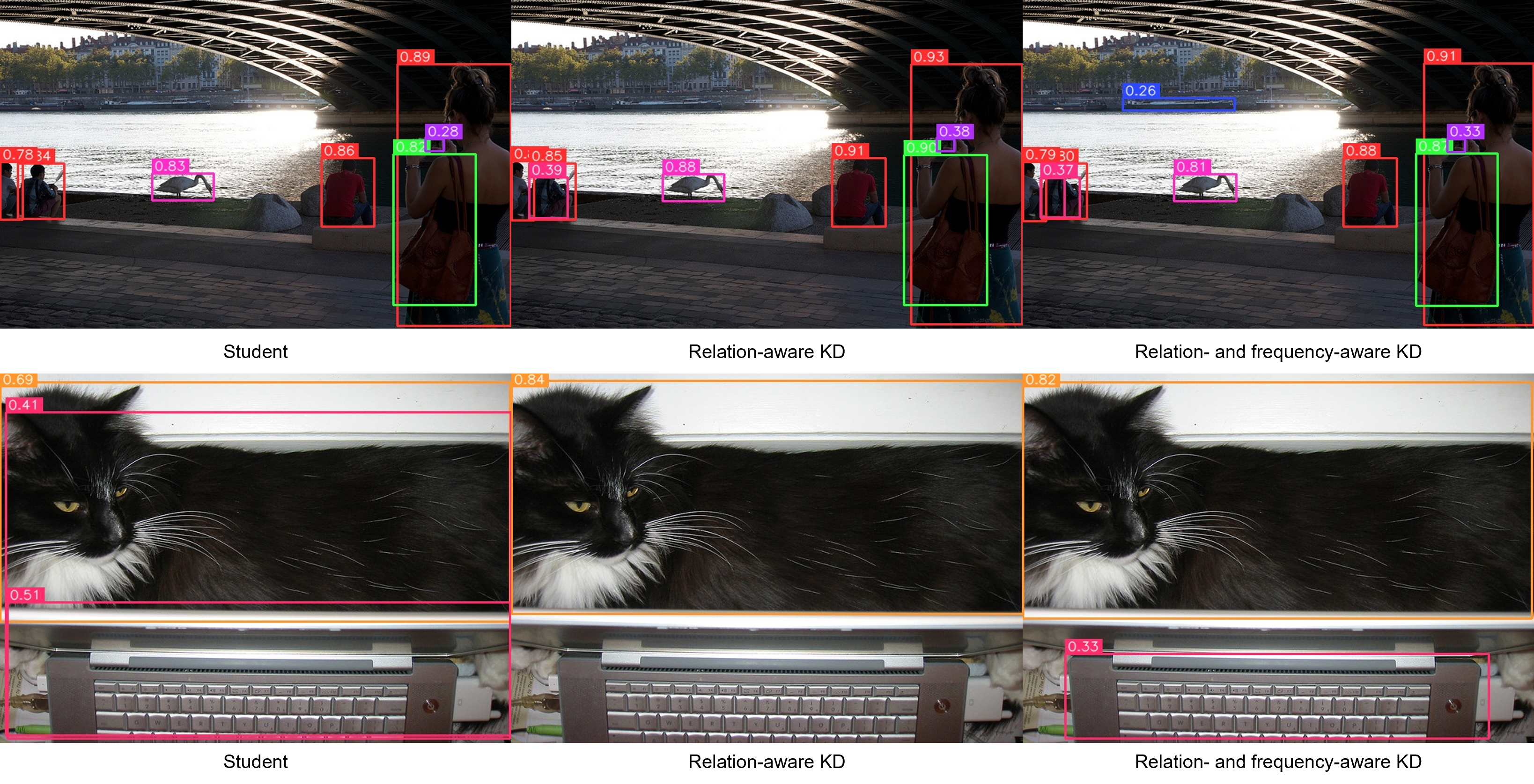}
  \caption{\textbf{Qualitative comparison on challenging scale and illumination variations.} From left to right in each group, the results correspond to the YOLOv12-M student, the relation-aware KD variant, and FD-CanKD. The relation-aware KD variant improves teacher--student spatial transfer, while FD-CanKD additionally applies different alignment rules to decomposed feature components. The examples illustrate more balanced detections under scale and illumination changes, especially for local objects that require fine spatial detail.}
  \label{fig:qualitative_scale_illumination}
\end{figure*}

\subsection{Scale-Pair Sensitivity Analysis}

Table~\ref{tab:scale_pair_generalization} evaluates FD-CanKD across different YOLOv12 teacher--student scale pairs under the from-scratch protocol.

In the YOLOv12-L \(\rightarrow\) YOLOv12-M setting, FD-CanKD reaches 46.4 \(\mathrm{mAP}_{50:95}\), compared with 45.6 for the student and 47.0 for the teacher. In the YOLOv12-X \(\rightarrow\) YOLOv12-M setting, FD-CanKD reaches 46.5, compared with 45.6 for the student and 49.5 for the teacher. In the YOLOv12-X \(\rightarrow\) YOLOv12-S setting, FD-CanKD reaches 41.1, compared with 40.9 for the student. These results show that the proposed distillation structure remains effective across the tested scale pairs, while the absolute AP depends on the teacher--student capacity relationship.

\subsection{Post-Distillation Fine-Tuning and Warm-Start Results}

Table~\ref{tab:finetune_warmstart} evaluates a post-training analysis rather than another from-scratch comparison. It compares detector-only YOLOv12-M fine-tuning, warm-start CanKD, and warm-start FD-CanKD, together with locally evaluated YOLOv12-M student and YOLOv12-X teacher references specific to this analysis. The Table~\ref{tab:finetune_warmstart} student reference obtains 45.63 \(\mathrm{mAP}_{50:95}\), 62.01 \(\mathrm{mAP}_{50}\), and 50.81 \(\mathrm{mAP}_{75}\), while the locally evaluated teacher obtains 49.58, 66.29, and 54.04, respectively. The same 45.63 local student reference is reused in Table~\ref{tab:object_size_analysis}, where its size-wise scores are 28.92 \(\mathrm{AP}_{S}\), 51.94 \(\mathrm{AP}_{M}\), and 62.35 \(\mathrm{AP}_{L}\). These post-training references are kept separate from the 45.6 diagnostic from-scratch student reference used in Tables~\ref{tab:external_kd_comparison}--\ref{tab:scale_pair_generalization}, so gains are interpreted within the corresponding evaluation protocol rather than across table groups.

Continuing detector-only fine-tuning reaches 47.02 \(\mathrm{mAP}_{50:95}\), 63.68 \(\mathrm{mAP}_{50}\), and 51.20 \(\mathrm{mAP}_{75}\) after 30 additional epochs. Under the same +30-epoch continued fine-tuning budget, CanKD reaches 48.80, 65.65, and 53.19, whereas FD-CanKD reaches 48.84, 65.82, and 53.35. Relative to detector-only fine-tuning at +30 epochs, FD-CanKD therefore improves the reported result by 1.82 \(\mathrm{mAP}_{50:95}\), 2.14 \(\mathrm{mAP}_{50}\), and 2.15 \(\mathrm{mAP}_{75}\). The strongest FD-CanKD checkpoint in Table~\ref{tab:finetune_warmstart} occurs at +20 epochs, reaching 48.87 \(\mathrm{mAP}_{50:95}\), 65.84 \(\mathrm{mAP}_{50}\), and 53.40 \(\mathrm{mAP}_{75}\). These results show that the observed refinement benefit is not explained by additional detector optimization alone. Teacher-guided relation transfer and frequency-aware alignment provide a more favorable representation for subsequent detector refinement.

To assess run-to-run variability at the distilled before-FT checkpoint, we repeated both CanKD and FD-CanKD with three random seeds. CanKD obtains 48.41 \(\pm\) 0.12 \(\mathrm{mAP}_{50:95}\), 65.09 \(\pm\) 0.13 \(\mathrm{mAP}_{50}\), and 52.79 \(\pm\) 0.18 \(\mathrm{mAP}_{75}\), whereas FD-CanKD obtains 48.42 \(\pm\) 0.07, 65.00 \(\pm\) 0.13, and 52.85 \(\pm\) 0.07, respectively. The before-FT mean difference in \(\mathrm{mAP}_{50:95}\) is therefore only 0.01 point, so the three-seed repetition is not used to claim a statistically significant before-FT superiority over CanKD. Instead, it serves as a reproducibility check: FD-CanKD reproduces a similar before-FT performance level across the tested seeds and shows lower observed sample standard deviation in \(\mathrm{mAP}_{50:95}\) and \(\mathrm{mAP}_{75}\). The practical refinement benefit is therefore supported primarily by the continued fine-tuning trajectory rather than by an overstated before-FT mean separation.

\subsection{Object-Size-Wise Analysis}

To identify where the frequency-decoupled branch contributes most clearly while avoiding dependence on a single selected checkpoint, Table~\ref{tab:object_size_analysis} first anchors the analysis with the same local YOLOv12-M student reference used in Table~\ref{tab:finetune_warmstart}: 45.63 \(\mathrm{mAP}_{50:95}\), 28.92 \(\mathrm{AP}_{S}\), 51.94 \(\mathrm{AP}_{M}\), and 62.35 \(\mathrm{AP}_{L}\). This row is a common reference only. The size-specific method comparison is made between CanKD and FD-CanKD at matched +20- and +30-epoch continued fine-tuning checkpoints. At +20 epochs, the overall \(\mathrm{mAP}_{50:95}\) difference between CanKD and FD-CanKD is modest at +0.12 point, whereas FD-CanKD improves \(\mathrm{AP}_{S}\) from 29.19 to 30.15, corresponding to a +0.96-point gain. The same direction remains at +30 epochs, where the overall gap narrows to +0.04 point but \(\mathrm{AP}_{S}\) still improves from 29.44 to 30.02, a +0.58-point gain. In contrast, \(\mathrm{AP}_{M}\) and \(\mathrm{AP}_{L}\) do not improve under FD-CanKD at either checkpoint. Therefore, the positive CanKD-to-FD-CanKD difference is repeatedly concentrated on small-object detection rather than uniformly distributed across all object scales.

This scale-specific behavior can be explained by how small objects are represented in multi-scale detection features. A small object occupies only a limited number of spatial locations, and repeated downsampling can substantially weaken its already sparse representation. Consequently, its remaining discriminative evidence is often concentrated in local intensity transitions, object boundaries, and fine structural details, which are more strongly associated with high-frequency responses than the broader structural patterns that dominate larger objects. Small-object recognition is therefore particularly sensitive to whether such fine-detail information survives feature transformation and teacher--student alignment.

% ---------------- Figure 5 ----------------
\begin{figure*}[!t]
  \centering
  \includegraphics[width=\textwidth]{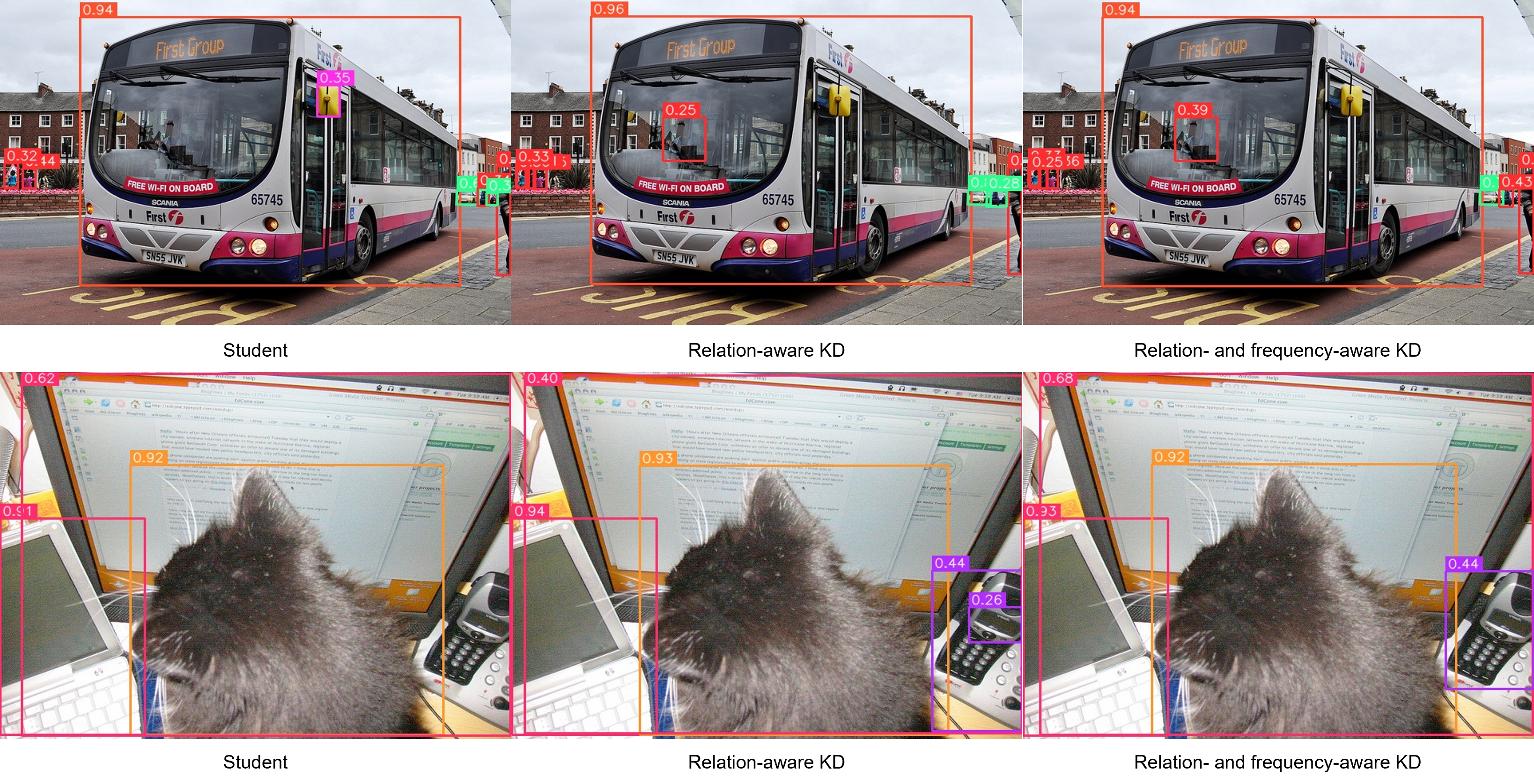}
  \caption{\textbf{Qualitative comparison on large objects and context-sensitive objects.} From left to right in each group, the results correspond to the YOLOv12-M student, the relation-aware KD variant, and FD-CanKD. The proposed method maintains reliable detection of dominant objects while improving the stability of context-sensitive and fine-grained object predictions in the illustrated examples.}
  \label{fig:qualitative_large_small}
\end{figure*}

The relation-aware transfer used in CanKD improves non-local contextual interaction by allowing the student to exploit teacher-side dependencies across spatial regions. However, relation transfer alone does not explicitly distinguish between slowly varying structural responses and rapidly varying detail-sensitive responses. Under a uniform alignment rule, dominant low-frequency structure may govern optimization, while weaker high-frequency responses associated with small boundaries or local details can receive comparatively limited emphasis. FD-CanKD instead decomposes the relation-enhanced representation into complementary frequency components and applies component-specific alignment objectives, preventing broad structural information and detail-sensitive responses from being forced to follow an identical matching rule.

This also explains why the improvement is not uniform across object sizes. Medium and large objects occupy more spatial locations and retain broader shape, regional, and contextual cues even after downsampling, whereas small objects have much less spatial redundancy retain broader shape, regional, and contextual cues even after down. The loss of only a few informative activations can therefore substantially weaken a small object's representation. The repeated \(AP_S\) advantage of FD-CanKD at both +20 and +30 epochs is thus consistent with the hypothesis that frequency-decoupled alignment helps preserve detail-sensitive teacher information that is disproportionately important for spatially small instances.

Importantly, this result should not be interpreted as evidence that all high-frequency information is beneficial or that the frequency branch alone causally determines small-object performance. High-frequency responses may also contain background clutter or noise, and Table~\ref{tab:object_size_analysis} provides a two-checkpoint size-wise diagnostic rather than a controlled causal isolation of individual frequency bands. We therefore interpret the repeated \(AP_S\) advantage as supportive quantitative evidence consistent with the proposed frequency-decoupled design, rather than as formal statistical proof of a causal frequency effect.

\subsection{Qualitative Results}

Figures~\ref{fig:qualitative_clutter},~\ref{fig:qualitative_scale_illumination}, and~\ref{fig:qualitative_large_small} present qualitative comparisons among the YOLOv12-M student, the relation-aware KD variant, and FD-CanKD. In each group, the results are arranged from left to right as the baseline student, relation-aware KD, and FD-CanKD. The highlighted boxes indicate representative changes in confidence, localization, or local object response across the compared methods. Although the aggregate numerical differences among KD variants are modest, the qualitative examples show that FD-CanKD more consistently preserves fine-grained local object responses and context-dependent detections in cluttered, occluded, or scale-varying scenes.

Figure~\ref{fig:qualitative_clutter} shows examples with cluttered scenes, overlapping objects, occlusion, and local objects. The baseline student detects dominant objects but often produces lower confidence predictions or unstable bounding boxes in cluttered regions. Relation-aware KD improves several responses by transferring teacher-side spatial context. FD-CanKD additionally applies band-specific alignment after relation transfer and shows more stable detections, including finer local object responses, in the illustrated cases. Because these examples are qualitative, they are used to complement rather than establish the interpretation of the decomposed feature components.

Figure~\ref{fig:qualitative_scale_illumination} compares results under challenging scale and illumination variations. In scenes containing distant persons, strong illumination contrast, and large foreground objects, the baseline student may miss context-sensitive objects. Relation-aware KD improves contextual object responses, while FD-CanKD additionally uses the asymmetric band-alignment objective and produces more balanced predictions for fine-grained local regions in the illustrated cases.

Figure~\ref{fig:qualitative_large_small} shows cases involving large objects and context-sensitive objects. The baseline student reliably detects large objects but can produce unstable responses for surrounding objects. Relation-aware KD improves some local responses through spatial context transfer. The final FD-CanKD model shows more consistent behavior across dominant objects and fine-grained local object regions.

These qualitative results complement the quantitative findings in Tables~\ref{tab:external_kd_comparison},~\ref{tab:yolov12_overall_comparison},~\ref{tab:finetune_warmstart}, and~\ref{tab:object_size_analysis}. While the aggregate numerical differences among from-scratch KD variants are small, the visual examples indicate that FD-CanKD better preserves fine-grained local object responses and context-dependent detections in challenging scenes. The object-size-wise breakdown provides additional quantitative support for this interpretation: FD-CanKD improves \(\mathrm{AP}_{S}\) over CanKD by 0.96 point at +20 epochs and by 0.58 point at +30 epochs, while gains are not uniformly observed for medium and large objects. The repeated direction across two neighboring checkpoints reduces dependence on a single selected refinement stage.

\subsection{Discussion}

The experimental results provide six observations. First, the from-scratch comparisons in Tables~\ref{tab:external_kd_comparison} and~\ref{tab:yolov12_overall_comparison} should be interpreted as
diagnostic evaluations under the fixed 50-epoch local protocol. FD-CanKD remains competitive with representative detector KD methods while using the locally reproduced YOLOv12-M student as a common
reference. Although the aggregate \(\mathrm{mAP}_{50:95}\) differences among the KD variants are small, FD-CanKD achieves the strongest \(\mathrm{mAP}_{50}\) in Table~\ref{tab:external_kd_comparison} and preserves the student-level \(\mathrm{mAP}_{75}\). This suggests that the proposed structure improves teacher-student alignment rather than merely adding more supervision terms.

Second, the qualitative results help explain why FD-CanKD remains useful despite modest aggregate AP differences. Figures~\ref{fig:qualitative_clutter}--\ref{fig:qualitative_large_small} show improved preservation of local object responses and context-dependent detections in cluttered, occluded, and scale-varying scenes. This behavior is consistent with relation-aware transfer capturing non-local spatial context and frequency-aware alignment preserving detail-sensitive cues for stable localization.

Third, Table~\ref{tab:object_size_analysis} provides object-size-wise evidence using the same 45.63 student reference as Table~\ref{tab:finetune_warmstart}. At +20 epochs, FD-CanKD improves
\(\mathrm{AP}_{S}\) from 29.19 to 30.15 over CanKD, a gain of 0.96 points, while the overall \(\mathrm{mAP}_{50:95}\) difference is 0.12 points. At +30 epochs, the overall difference decreases to
0.04 points, but the \(\mathrm{AP}_{S}\) gain remains 0.58 points, from 29.44 to 30.02. Comparable improvements are not observed for medium or large objects. The repeated concentration of the positive
difference in \(\mathrm{AP}_{S}\) is consistent with the motivation that frequency-aware alignment benefits boundary and fine-detail responses, although this analysis is treated as diagnostic evidence
rather than a formal significance test.

Fourth, Table~\ref{tab:finetune_warmstart} demonstrates the benefit of KD-guided refinement over detector-only continued fine-tuning. After 30 additional epochs, detector-only fine-tuning reaches 47.02
\(\mathrm{mAP}_{50:95}\), 63.68 \(\mathrm{mAP}_{50}\), and 51.20 \(\mathrm{mAP}_{75}\). Under the same budget, CanKD reaches 48.80, 65.65, and 53.19, whereas FD-CanKD reaches 48.84, 65.82, and 53.35.
FD-CanKD obtains its strongest checkpoint after 20 additional epochs, reaching 48.87, 65.84, and 53.40, respectively. These results show that teacher-guided refinement provides a more favorable optimization
trajectory than detector-only fine-tuning.

Fifth, the three-seed experiment confirms reproducibility at the distilled before-FT checkpoint. CanKD obtains 48.41 \(\pm\) 0.12 \(\mathrm{mAP}_{50:95}\), whereas FD-CanKD obtains 48.42 \(\pm\) 0.07. The 0.01-point difference is negligible and is not interpreted as statistical superiority. Instead, the lower
observed variation for FD-CanKD supports that its before-FT result is reproducible across the tested seeds rather than being caused by one favorable run.

Sixth, the frequency and scale-pair ablations clarify the source and scope of the observed improvements. Component-selective frequency alignment helps preserve stricter localization performance, while the
scale-pair experiments show that FD-CanKD remains effective across the evaluated teacher-student sizes under the same local protocol. Together, these findings support FD-CanKD as a structural distillation framework that prepares compact detectors for subsequent refinement.

We note that all reported results are obtained within the YOLOv12 teacher--student family. This single-family control was a deliberate design choice for isolating component effects, and validating the proposed alignment principles on other detector families (e.g., DETR-style or other YOLO variants) is left for future work. 

Overall, FD-CanKD should be interpreted as a detector-oriented teacher-student alignment framework rather than an accumulation of auxiliary objectives. It organizes teacher knowledge into prediction-level, relation-level, and frequency-aware cues and applies an appropriate alignment rule to each type. Because all distillation components are removed after training, the deployed YOLOv12-M model retains its original architecture and parameter count. The primary benefit is therefore improved accuracy and refinement behavior without additional inference complexity.

\section{Conclusion}

In this paper, we presented FD-CanKD, a detector-oriented integration framework that combines head-level prediction supervision, cross-attention-based relation transfer, and frequency-decoupled alignment for controlled YOLOv12 large-to-small distillation. Unlike single-target feature matching, FD-CanKD organizes teacher knowledge into complementary prediction, relation, and frequency-aware cues and transfers them with alignment rules suited to dense object detection.

The experiments show that the locally reproduced YOLOv12-M student obtains 45.6 \(\mathrm{mAP}_{50:95}\) as the controlled 50-epoch from-scratch reference for the diagnostic comparisons in Tables~\ref{tab:external_kd_comparison}--\ref{tab:scale_pair_generalization}. Under this setting, representative detector KD methods and FD-CanKD achieve similar aggregate \(\mathrm{mAP}_{50:95}\) values, while FD-CanKD provides the strongest \(\mathrm{mAP}_{50}\) among the compared KD variants and preserves the student-level \(\mathrm{mAP}_{75}\). Qualitative results further show that FD-CanKD better preserves fine-grained local object responses in cluttered, occluded, and scale-varying scenes. In the separate post-distillation continued fine-tuning analysis of Table~\ref{tab:finetune_warmstart}, detector-only YOLOv12-M fine-tuning reaches 47.02 \(\mathrm{mAP}_{50:95}\), 63.68 \(\mathrm{mAP}_{50}\), and 51.20 \(\mathrm{mAP}_{75}\) after 30 additional epochs, while FD-CanKD reaches 48.84, 65.82, and 53.35 under the same +30-epoch budget and peaks at 48.87, 65.84, and 53.40 after 20 additional epochs. At the distilled before-FT checkpoint, three-seed repetition gives 48.41 \(\pm\) 0.12 \(\mathrm{mAP}_{50:95}\) for CanKD and 48.42 \(\pm\) 0.07 for FD-CanKD. Because the mean difference is only 0.01 point, these seed repetitions are interpreted as a reproducibility check rather than as evidence of a statistically significant before-FT advantage. Across the shared +20- and +30-epoch checkpoints, the object-size-wise breakdown shows that the positive difference is consistently concentrated in small-object detection: FD-CanKD improves \(\mathrm{AP}_{S}\) from 29.19 to 30.15 over CanKD (+0.96 point) at +20 epochs and from 29.44 to 30.02 (+0.58 point) at +30 epochs. This repeated trend provides quantitative evidence consistent with the proposed fine-detail alignment motivation while avoiding reliance on a single selected checkpoint.

These results indicate that relation-aware and frequency-aware distillation provides a more favorable refinement starting point than detector-only fine-tuning for compact object detectors. Since all auxiliary distillation modules are removed after training, the deployed model remains the original 19.7M-parameter YOLOv12-M student. All experiments in this study are conducted within a single representative teacher--student family, YOLOv12. This setting was deliberately controlled to isolate the effect of relation- and frequency-aware alignment without confounding differences in codebase, training schedule, or initialization across detector architectures. Extending FD-CanKD to other detector families remains an important direction for future work, together with adaptive distillation scheduling, instance-aware alignment, and uncertainty-guided feature selection for more robust compact detector optimization.

\bmsubsection*{Use of Generative AI}
During the preparation of this manuscript, the authors used OpenAI's ChatGPT to assist with English-language editing, sentence restructuring, and improving the clarity and readability of the manuscript. All AI-assisted content was critically reviewed and revised by the authors, who take full responsibility for the final content of the manuscript.

\bmsubsection*{Author Contributions}
YoungJae Cheong: Conceptualization, Methodology, Software, Validation, Formal Analysis, Investigation, Visualization, and Writing -- Original Draft. Jhonghyun An: Conceptualization, Methodology, Supervision, Project Administration, Funding Acquisition, and Writing -- Review \& Editing. All authors have read and approved the final manuscript.

\bmsubsection*{Acknowledgements}
This work was supported by the Institute of Information \& Communications Technology Planning \& Evaluation (IITP) grant funded by the Ministry of Science and ICT (MSIT), Republic of Korea (No. RS-2024-00397615, ``Development of SDV-Based Automotive Software Platform Technology Interacted with an AI Framework Required for Intelligent Vehicles'').

\bmsubsection*{Financial Disclosure}
The authors declare no relevant financial interests to disclose.

\bmsubsection*{Conflicts of Interest}
The authors declare no conflicts of interest.

\bmsubsection*{Data Availability Statement}
This study uses the publicly available Microsoft COCO dataset for training and evaluation. The dataset can be accessed through the official COCO repository. No additional dataset was generated as part of this work.

% ---------------- References ----------------

\bmsubsection*{Code Availability Statement}
The implementation code, training configuration, hyperparameter settings, and evaluation scripts can be made available from the corresponding author upon reasonable request.


\begin{thebibliography}{99}

\bibitem{redmon2016yolo}
J. Redmon, S. Divvala, R. Girshick, and A. Farhadi, ``You only look once: Unified, real-time object detection,'' in \emph{Proc. IEEE Conf. Comput. Vis. Pattern Recognit. (CVPR)}, 2016, pp. 779--788.

\bibitem{redmon2017yolo9000}
J. Redmon and A. Farhadi, ``YOLO9000: Better, faster, stronger,'' in \emph{Proc. IEEE Conf. Comput. Vis. Pattern Recognit. (CVPR)}, 2017, pp. 7263--7271.

\bibitem{redmon2018yolov3}
J. Redmon and A. Farhadi, ``YOLOv3: An incremental improvement,'' \emph{arXiv preprint arXiv:1804.02767}, 2018.

\bibitem{bochkovskiy2020yolov4}
A. Bochkovskiy, C.-Y. Wang, and H.-Y. M. Liao, ``YOLOv4: Optimal speed and accuracy of object detection,'' \emph{arXiv preprint arXiv:2004.10934}, 2020.

\bibitem{jocher2020yolov5}
G. Jocher, K. Nishimura, T. Mineeva, and R. J. A. M. Vilariño, ``YOLOv5,'' 2020. [Online]. Available: \url{https://github.com/ultralytics/yolov5}

\bibitem{li2023yolov6}
C. Li, L. Li, Y. Geng, H. Jiang, M. Cheng, B. Zhang, Z. Ke, X. Xu, and X. Chu, ``YOLOv6 v3.0: A full-scale reloading,'' \emph{arXiv preprint arXiv:2301.05586}, 2023.

\bibitem{wang2023yolov7}
C.-Y. Wang, A. Bochkovskiy, and H.-Y. M. Liao, ``YOLOv7: Trainable bag-of-freebies sets new state-of-the-art for real-time object detectors,'' in \emph{Proc. IEEE/CVF Conf. Comput. Vis. Pattern Recognit. (CVPR)}, 2023, pp. 7464--7475.

\bibitem{jocher2023yolov8}
Ultralytics, ``YOLOv8 anchor-free bounding box prediction,'' 2023. [Online]. Available: \url{https://github.com/ultralytics/ultralytics/issues/189}

\bibitem{wang2024yolov9}
C.-Y. Wang, I.-H. Yeh, and H.-Y. M. Liao, ``YOLOv9: Learning what you want to learn using programmable gradient information,'' in \emph{Proc. Eur. Conf. Comput. Vis. (ECCV)}, 2024, pp. 1--21.

\bibitem{wang2024yolov10}
A. Wang, H. Chen, L. Liu, K. Chen, Z. Lin, J. Han, and G. Ding, ``YOLOv10: Real-time end-to-end object detection,'' \emph{Adv. Neural Inf. Process. Syst.}, vol. 37, pp. 107984--108011, 2024.

\bibitem{jocher2024yolov11}
G. Jocher, ``YOLOv11,'' 2024. [Online]. Available: \url{https://github.com/ultralytics}

\bibitem{tian2025yolov12}
Y. Tian, Q. Ye, and D. Doermann, ``YOLOv12: Attention-centric real-time object detectors,'' in \emph{Advances in Neural Information Processing Systems (NeurIPS)}, vol. 38, 2025.

\bibitem{li2022dndetr}
F. Li, H. Zhang, S. Liu, J. Guo, L. M. Ni, and L. Zhang, ``DN-DETR: Accelerate DETR training by introducing query denoising,'' in \emph{Proc. IEEE/CVF Conf. Comput. Vis. Pattern Recognit. (CVPR)}, 2022, pp. 13619--13627.

\bibitem{zhao2024rtdetr}
Y. Zhao, W. Lv, S. Xu, J. Wei, G. Wang, Q. Dang, Y. Liu, and J. Chen, ``DETRs beat YOLOs on real-time object detection,'' in \emph{Proc. IEEE/CVF Conf. Comput. Vis. Pattern Recognit. (CVPR)}, 2024, pp. 16965--16974.

\bibitem{lv2024rtdetrv2}
W. Lv, Y. Zhao, Q. Chang, K. Huang, G. Wang, and Y. Liu, ``RT-DETRv2: Improved baseline with bag-of-freebies for efficient detection transformer,'' \emph{arXiv preprint arXiv:2407.17140}, 2024.

\bibitem{bucilua2006model}
C. Bucilua, R. Caruana, and A. Niculescu-Mizil, ``Model compression,'' in \emph{Proc. 12th ACM SIGKDD Int. Conf. Knowl. Discov. Data Min. (KDD)}, 2006, pp. 535--541.

\bibitem{hinton2015distilling}
G. Hinton, O. Vinyals, and J. Dean, ``Distilling the knowledge in a neural network,'' \emph{arXiv preprint arXiv:1503.02531}, 2015.

\bibitem{romero2014fitnets}
A. Romero, N. Ballas, S. E. Kahou, A. Chassang, C. Gatta, and Y. Bengio, ``FitNets: Hints for thin deep nets,'' \emph{arXiv preprint arXiv:1412.6550}, 2014.

\bibitem{zhang2018dml}
Y. Zhang, T. Xiang, T. M. Hospedales, and H. Lu, ``Deep mutual learning,'' in \emph{Proc. IEEE Conf. Comput. Vis. Pattern Recognit. (CVPR)}, 2018, pp. 4320--4328.

\bibitem{zhao2022dkd}
B. Zhao, Q. Cui, R. Song, Y. Qiu, and J. Liang, ``Decoupled knowledge distillation,'' in \emph{Proc. IEEE/CVF Conf. Comput. Vis. Pattern Recognit. (CVPR)}, 2022, pp. 11953--11962.

\bibitem{huang2022dist}
T. Huang, S. You, F. Wang, C. Qian, and C. Xu, ``Knowledge distillation from a stronger teacher,'' \emph{Adv. Neural Inf. Process. Syst.}, vol. 35, pp. 33716--33727, 2022.

\bibitem{chen2021review}
P. Chen, S. Liu, H. Zhao, and J. Jia, ``Distilling knowledge via knowledge review,'' in \emph{Proc. IEEE/CVF Conf. Comput. Vis. Pattern Recognit. (CVPR)}, 2021, pp. 5008--5017.

\bibitem{park2019rkd}
W. Park, D. Kim, Y. Lu, and M. Cho, ``Relational knowledge distillation,'' in \emph{Proc. IEEE/CVF Conf. Comput. Vis. Pattern Recognit. (CVPR)}, 2019, pp. 3967--3976.

\bibitem{cao2022pkd}
W. Cao, Y. Zhang, J. Gao, A. Cheng, K. Cheng, and J. Cheng, ``PKD: General distillation framework for object detectors via Pearson correlation coefficient,'' \emph{Adv. Neural Inf. Process. Syst.}, vol. 35, pp. 15394--15406, 2022.

\bibitem{heo2019overhaul}
B. Heo, J. Kim, S. Yun, H. Park, N. Kwak, and J. Y. Choi, ``A comprehensive overhaul of feature distillation,'' in \emph{Proc. IEEE/CVF Int. Conf. Comput. Vis. (ICCV)}, 2019, pp. 1921--1930.

\bibitem{li2017mimic}
Q. Li, S. Jin, and J. Yan, ``Mimicking very efficient network for object detection,'' in \emph{Proc. IEEE Conf. Comput. Vis. Pattern Recognit. (CVPR)}, 2017, pp. 6356--6364.

\bibitem{wang2021mimicking}
G.-H. Wang, Y. Ge, and J. Wu, ``Distilling knowledge by mimicking features,'' \emph{IEEE Trans. Pattern Anal. Mach. Intell.}, vol. 44, no. 11, pp. 8183--8195, 2021.

\bibitem{zhang2020fkd}
L. Zhang and K. Ma, ``Improve object detection with feature-based knowledge distillation: Towards accurate and efficient detectors,'' in \emph{Proc. Int. Conf. Learn. Represent. (ICLR)}, 2020.

\bibitem{yang2022fgd}
Z. Yang, Z. Li, X. Jiang, Y. Gong, Z. Yuan, D. Zhao, and C. Yuan, ``Focal and global knowledge distillation for detectors,'' in \emph{Proc. IEEE/CVF Conf. Comput. Vis. Pattern Recognit. (CVPR)}, 2022, pp. 4643--4652.

\bibitem{shu2021cwd}
C. Shu, Y. Liu, J. Gao, Z. Yan, and C. Shen, ``Channel-wise knowledge distillation for dense prediction,'' in \emph{Proc. IEEE/CVF Int. Conf. Comput. Vis. (ICCV)}, 2021, pp. 5311--5320.

\bibitem{yang2022mgd}
Z. Yang, Z. Li, M. Shao, D. Shi, Z. Yuan, and C. Yuan, ``Masked generative distillation,'' in \emph{Proc. Eur. Conf. Comput. Vis. (ECCV)}, 2022, pp. 53--69.

\bibitem{huang2023diffkd}
T. Huang, Y. Zhang, M. Zheng, S. You, F. Wang, C. Qian, and C. Xu, ``Knowledge diffusion for distillation,'' \emph{Adv. Neural Inf. Process. Syst.}, vol. 36, pp. 65299--65316, 2023.

\bibitem{kang2021icd}
Z. Kang, P. Zhang, X. Zhang, J. Sun, and N. Zheng, ``Instance-conditional knowledge distillation for object detection,'' \emph{Adv. Neural Inf. Process. Syst.}, vol. 34, pp. 16468--16480, 2021.

\bibitem{li2024detkds}
L. Li, Y. Bao, P. Dong, C. Yang, A. Li, W. Luo, Q. Liu, W. Xue, and Y. Guo, ``DetKDS: Knowledge distillation search for object detectors,'' in \emph{Proc. Int. Conf. Mach. Learn. (ICML)}, 2024.

\bibitem{zhou2024pcka}
Z. Zhou, Y. Shen, S. Shao, L. Gong, and S. Lin, ``Rethinking centered kernel alignment in knowledge distillation,'' \emph{arXiv preprint arXiv:2401.11824}, 2024.

\bibitem{zhang2024freekd}
Y. Zhang, T. Huang, J. Liu, T. Jiang, K. Cheng, and S. Zhang, ``FreeKD: Knowledge distillation via semantic frequency prompt,'' in \emph{Proc. IEEE/CVF Conf. Comput. Vis. Pattern Recognit. (CVPR)}, 2024, pp. 15931--15940.

\bibitem{vaswani2017attention}
A. Vaswani, N. Shazeer, N. Parmar, J. Uszkoreit, L. Jones, A. N. Gomez, L. Kaiser, and I. Polosukhin, ``Attention is all you need,'' \emph{Adv. Neural Inf. Process. Syst.}, vol. 30, 2017.

\bibitem{wang2018nonlocal}
X. Wang, R. Girshick, A. Gupta, and K. He, ``Non-local neural networks,'' in \emph{Proc. IEEE Conf. Comput. Vis. Pattern Recognit. (CVPR)}, 2018, pp. 7794--7803.

\bibitem{hu2018relation}
H. Hu, J. Gu, Z. Zhang, J. Dai, and Y. Wei, ``Relation networks for object detection,'' in \emph{Proc. IEEE Conf. Comput. Vis. Pattern Recognit. (CVPR)}, 2018, pp. 3588--3597.

\bibitem{sun2025cankd}
S. Sun and W. Ohyama, ``CanKD: Cross-attention-based non-local operation for feature-based knowledge distillation,'' in \emph{Proc. IEEE/CVF Winter Conf. Appl. Comput. Vis. (WACV)}, 2026, pp. 8606--8616.

\bibitem{huo2024c2kd}
F. Huo, W. Xu, J. Guo, H. Wang, and S. Guo, ``C2KD: Bridging the modality gap for cross-modal knowledge distillation,'' in \emph{Proc. IEEE/CVF Conf. Comput. Vis. Pattern Recognit. (CVPR)}, 2024, pp. 16006--16015.

\bibitem{liu2023emotionkd}
Y. Liu, Z. Jia, and H. Wang, ``EmotionKD: A cross-modal knowledge distillation framework for emotion recognition based on physiological signals,'' in \emph{Proc. 31st ACM Int. Conf. Multimedia (ACM MM)}, 2023, pp. 6122--6131.

\bibitem{sun2016deepcoral}
B. Sun and K. Saenko, ``Deep CORAL: Correlation alignment for deep domain adaptation,'' in \emph{Proc. Eur. Conf. Comput. Vis. Workshops (ECCVW)}, 2016, pp. 443--450.

\bibitem{bruna2013scattering}
J. Bruna and S. Mallat, ``Invariant scattering convolution networks,'' \emph{IEEE Trans. Pattern Anal. Mach. Intell.}, vol. 35, no. 8, pp. 1872--1886, 2013.

\bibitem{williams2018wavelet}
T. Williams and R. Li, ``Wavelet pooling for convolutional neural networks,'' in \emph{Proc. Int. Conf. Learn. Represent. (ICLR)}, 2018.

\bibitem{xu2020frequency}
K. Xu, M. Qin, F. Sun, Y. Wang, Y.-K. Chen, and F. Ren, ``Learning in the frequency domain,'' in \emph{Proc. IEEE/CVF Conf. Comput. Vis. Pattern Recognit. (CVPR)}, 2020, pp. 1740--1749.

\bibitem{pham2024frequency}
C. Pham, V.-A. Nguyen, T. Le, D. Phung, G. Carneiro, and T.-T. Do, ``Frequency attention for knowledge distillation,'' in \emph{Proc. IEEE/CVF Winter Conf. Appl. Comput. Vis. (WACV)}, 2024, pp. 2277--2286.

\bibitem{lin2014coco}
T.-Y. Lin, M. Maire, S. Belongie, J. Hays, P. Perona, D. Ramanan, P. Dollár, and C. L. Zitnick, ``Microsoft COCO: Common objects in context,'' in \emph{Proc. Eur. Conf. Comput. Vis. (ECCV)}, 2014, pp. 740--755.

\bibitem{liu2025fdcmkd}
J. Liu, Y. Zhang, T. Huang, W. Xu, and R. Yang, ``Distilling cross-modal knowledge via feature disentanglement,'' in \emph{Proc. AAAI Conf. Artif. Intell.}, vol. 40, no. 28, 2026, pp. 23739--23747, doi: 10.1609/aaai.v40i28.39548.

\bibitem{wang2020cspnet}
C.-Y. Wang, H.-Y. M. Liao, Y.-H. Wu, P.-Y. Chen, J.-W. Hsieh, and I.-H. Yeh, ``CSPNet: A new backbone that can enhance learning capability of CNN,'' in \emph{Proc. IEEE/CVF Conf. Comput. Vis. Pattern Recognit. Workshops (CVPRW)}, 2020, pp. 390--391.

\bibitem{wang2022elan}
C.-Y. Wang, H.-Y. M. Liao, and I.-H. Yeh, ``Designing network design strategies through gradient path analysis,'' \emph{arXiv preprint arXiv:2211.04800}, 2022.

\bibitem{dao2022flashattention}
T. Dao, D. Fu, S. Ermon, A. Rudra, and C. Ré, ``FlashAttention: Fast and memory-efficient exact attention with IO-awareness,'' \emph{Adv. Neural Inf. Process. Syst.}, vol. 35, pp. 16344--16359, 2022.

\bibitem{dao2023flashattention2}
T. Dao, ``FlashAttention-2: Faster attention with better parallelism and work partitioning,'' \emph{arXiv preprint arXiv:2307.08691}, 2023.

\bibitem{lin2017fpn}
T.-Y. Lin, P. Dollár, R. Girshick, K. He, B. Hariharan, and S. Belongie, ``Feature pyramid networks for object detection,'' in \emph{Proc. IEEE Conf. Comput. Vis. Pattern Recognit. (CVPR)}, 2017, pp. 2117--2125.

\bibitem{rezatofighi2019giou}
H. Rezatofighi, N. Tsoi, J. Gwak, A. Sadeghian, I. Reid, and S. Savarese, ``Generalized intersection over union: A metric and a loss for bounding box regression,'' in \emph{Proc. IEEE/CVF Conf. Comput. Vis. Pattern Recognit. (CVPR)}, 2019, pp. 658--666.

\bibitem{zheng2020diou}
Z. Zheng, P. Wang, W. Liu, J. Li, R. Ye, and D. Ren, ``Distance-IoU loss: Faster and better learning for bounding box regression,'' in \emph{Proc. AAAI Conf. Artif. Intell.}, vol. 34, no. 7, 2020, pp. 12993--13000.

\bibitem{zhou2019iouloss}
D. Zhou, J. Fang, X. Song, C. Guan, J. Yin, Y. Dai, and R. Yang, ``IoU loss for 2D/3D object detection,'' in \emph{Proc. Int. Conf. 3D Vis. (3DV)}, 2019, pp. 85--94.

\bibitem{li2020gfl}
X. Li, W. Wang, L. Wu, S. Chen, X. Hu, J. Li, J. Tang, and J. Yang, ``Generalized focal loss: Learning qualified and distributed bounding boxes for dense object detection,'' \emph{Adv. Neural Inf. Process. Syst.}, vol. 33, pp. 21002--21012, 2020.

\end{thebibliography}
\end{document}